\documentclass[lettersize,journal]{IEEEtran}

\usepackage{booktabs}       \usepackage{subcaption}
\usepackage{amsfonts}       \usepackage{nicefrac}       \usepackage{microtype}      \usepackage{xcolor}         \usepackage{colortbl}

\usepackage{times}
\usepackage{epsfig}
\usepackage{amssymb}
\usepackage{mathtools}
\usepackage{enumitem}
\usepackage{setspace}
\usepackage{booktabs}
\usepackage{color,xcolor}
\usepackage{bbding}
\usepackage{diagbox}
\usepackage{bm}
\usepackage{multirow}
\usepackage{caption}
\usepackage{makecell}
\usepackage{afterpage}
\usepackage{placeins}
\usepackage{float}
\usepackage{xspace}
\usepackage{ragged2e}
\usepackage{amsmath}
\usepackage{physics} \usepackage{url}
\usepackage[numbers,compress,sort]{natbib}

\newcommand{\nickname}{\mbox{ViVLA}\xspace}
\newcommand{\dataname}{\mbox{Human2Robot}\xspace}

\newcommand{\mycolor}{black}
\newcommand{\myindent}{}

\begin{document}

\title{See Once, Then Act: Vision-Language-Action Model with \\ Task Learning from One-Shot Video Demonstrations}

\author{Guangyan Chen$^{1\dagger}$, Meiling Wang$^{1}$, Qi Shao$^{1}$, Zichen Zhou$^{1}$, Weixin Mao$^{2}$, Te Cui$^{1}$, Minzhao Zhu$^{2}$, Yinan Deng$^{1}$, Luojie Yang$^{1}$, Zhanqi Zhang$^{2}$, Yi Yang${^{1}}$, Hua Chen$^{2}$, Yufeng Yue${^{1*}}$ 
\thanks{$^{*}$Corresponding author. 

$^\dagger$Guangyan Chen completed this work during an internship at LimX Dynamics.} \\
  $^1$ Beijing Institute of Technology \quad
  $^2$ LimX Dynamics    
}

\maketitle

\begin{abstract}
Developing robust and general-purpose manipulation policies represents a fundamental objective in robotics research.
While Vision-Language-Action (VLA) models have demonstrated promising capabilities for end-to-end robot control, {existing approaches still exhibit limited generalization to tasks beyond their training distributions.}
\textcolor{\mycolor}{In contrast, humans possess remarkable proficiency in acquiring novel skills by simply observing others performing them once.
Inspired by this capability, we propose \nickname, a generalist robotic manipulation policy that achieves efficient task learning from a single expert demonstration video at test time.} 
Our approach jointly processes an expert demonstration video alongside the robot's visual observations to predict both the demonstrated action sequences and subsequent robot actions, effectively distilling fine-grained manipulation knowledge from expert behavior and transferring it seamlessly to the agent.
\textcolor{\mycolor}{To enhance the performance of \nickname, we develop a scalable expert-agent pair data generation pipeline capable of synthesizing paired trajectories from easily accessible human videos, further augmented by curated pairs from publicly available datasets. This pipeline produces a total of 892,911 expert-agent samples for training \nickname.
Experimental results demonstrate that our \nickname is able to acquire novel manipulation skills from only a single expert demonstration video at test time.}
Our approach achieves over 30\% improvement on unseen LIBERO tasks and maintains above 35\% gains with cross-embodiment videos. Real-world experiments demonstrate effective learning from human videos, yielding more than 38\% improvement on unseen tasks.

\end{abstract}

\begin{IEEEkeywords}
 One-Shot Visual Imitation Learning, Vision Language Action Models, Unseen Task Generalization, Cross-embodiment Transfer, Robot Policy Learning.
\end{IEEEkeywords}

\begin{figure*}[t]
    \begin{center}
    \includegraphics[width=0.98\textwidth]{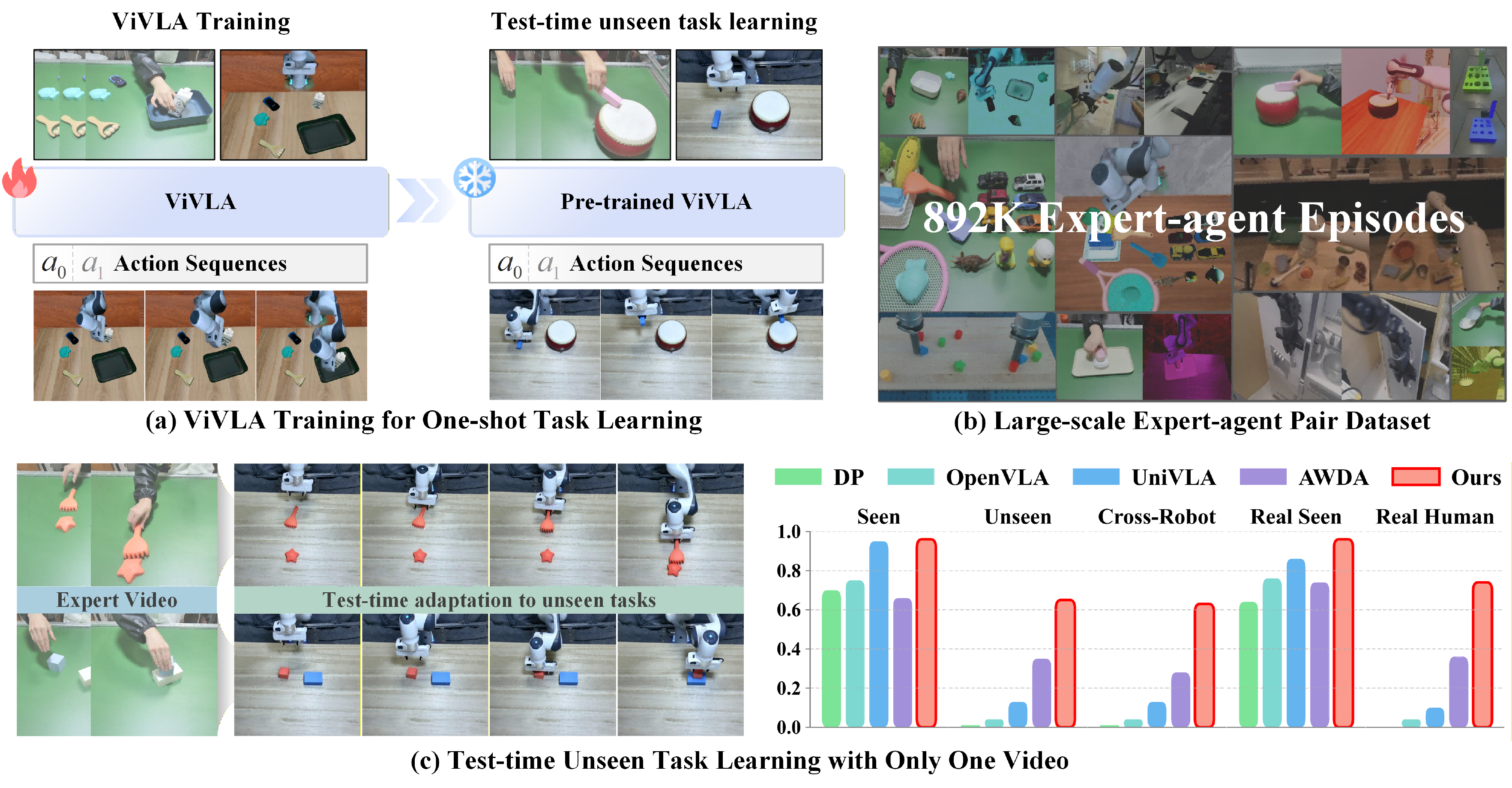}
           \vspace{-0.cm}
               \captionof{figure}{{Illustration of our \nickname. (a) Our \nickname is trained to predict subsequent robotic actions conditioned on a single expert demonstration video, endowing the model with the capacity to learn novel tasks from a single expert demonstration video at test time. (b) To push the performance limit of our proposed \nickname, we develop a scalable expert-agent pair data generation pipeline and compile a large-scale expert-agent pair dataset. (c) Extensive experiments demonstrate that our proposed \nickname efficiently learns unseen tasks and achieves state-of-the-art performance.}}
    \label{Fig::Figure1}
        \end{center}
        \vspace{-10pt}
\end{figure*}

\section{Introduction}
\label{sec:intro}

A fundamental objective in robotics research is the development of versatile, general-purpose robotic systems capable of performing diverse tasks across multiple domains, approaching human-level adaptability and versatility.
Inspired by the remarkable success of Large Language Models (LLMs) \cite{liu2024deepseek, bai2023qwen, touvron2023llama, achiam2023gpt} and Vision Language Models (VLMs) \cite{bai2025qwen2, team2025kimi, team2024chameleon, liu2023visual, xiong2025bluelm}, which are capable of generalizing across a wide range of tasks and domains, the robotics community is actively developing robotic foundation models with similar capabilities.
These works \cite{kim2024openvla, black2024pi_0, intelligence2025pi_, qu2025spatialvla, fan2025interleave, bu2025univla} typically build upon pretrained vision-language models and perform subsequent training on large-scale robot datasets spanning diverse objects, environments, and skills, thereby endowing robotic systems with extensive manipulation knowledge and enhanced generalization capabilities.
Despite these advances, generalizing to tasks beyond their training distributions remains a significant challenge. 
In stark contrast, humans exhibit remarkably efficient learning through visual imitation, extracting task-relevant knowledge from expert demonstrations and reproducing similar behaviors to accomplish comparable objectives.
Inspired by this desirable capability, a natural question arises: \textit{Can robotic agents similarly learn novel manipulation tasks directly by observing a single expert demonstration video?}

To answer this question, we investigate the development of a generalist policy for robotics that enables the robot to learn novel tasks beyond its training distribution by observing a single expert demonstration video without additional training. We develop our model based on VLMs to leverage their extensive prior knowledge and video comprehension capabilities. Upon this foundation, we train the model to predict subsequent robotic actions conditioned on a single expert demonstration video, endowing the model with the capacity to learn novel tasks from a single expert demonstration video at test time.
While this approach is conceptually promising and benefits from powerful VLMs, achieving such capabilities remains non-trivial due to the following reasons:

(\uppercase\expandafter{\romannumeral1})
\textbf{Model capability: Lack of fine-grained action recognition capability.}
A critical capability for enabling such generalist robotic policies is the acquisition of fine-grained manipulation knowledge from expert videos. However, existing VLMs predominantly focus on semantic-level video comprehension and exhibit limited proficiency in discerning fine-grained manipulation actions within video sequences.
To address these limitations, we introduce a fine-grained action reasoning objective during training, wherein the model is trained to explicitly articulate the manipulation actions observed in the demonstration video prior to generating corresponding robotic control outputs, thereby enhancing its capacity for fine-grained action recognition and comprehension. 
Furthermore, we incorporate a temporal localization task that inserts the agent's observation images into the demonstration video sequence, training the model to identify the temporal position of these observations within the video, thereby facilitating cross-modal information exchange between video and image representations.

(\uppercase\expandafter{\romannumeral2})
\textbf{Action representation: Lack of action labels for video data and the discrepancy in action spaces between the agent and the expert.}
Video data typically lacks action annotations, particularly in human videos, impeding the training of VLMs on fine-grained actions. Furthermore, experts in demonstration videos and target agents typically involve different embodiments, and the resulting disparity in action spaces complicates effective knowledge transfer from expert demonstration videos to the agent robot. 
To address the absence of action annotations, we propose training a latent action tokenizer that derives action representations directly from visual observations. To further bridge the embodiment gap, we train this tokenizer jointly on both expert demonstration videos and agent trajectory data, and introduce an action-centric cycle consistency objective to regularize the learned latent action space, establishing a unified action representation across the agent and the expert.
Following the training of the latent action tokenizer, VLMs are trained to predict latent actions from both expert videos and agent observations within this unified latent action space, enabling effective knowledge transfer from expert demonstrations to robotic agents.

(\uppercase\expandafter{\romannumeral3}) 
\textbf{Action modeling strategy: Autoregressive action modeling strategy results in shortcut learning and increased inference latency.}
In autoregressive next-action prediction training, the model has access to all preceding ground-truth action tokens within the sequence. This accessibility allows the model to exploit preceding ground-truth action tokens to predict subsequent actions, \textcolor{\mycolor}{thereby inducing shortcut learning and hindering the development of a genuine understanding of expert demonstration videos and agent observations.} Moreover, the autoregressive modeling paradigm necessitates sequential token-by-token generation, which introduces substantial inference latency.
To mitigate these limitations, we adopt a parallel decoding strategy where the model receives empty action embeddings as input and generates all action tokens concurrently in a single forward pass. This modification prevents information leakage from preceding action sequences, thereby compelling the model to ground its predictions in a comprehensive analysis of video content and agent observations. Furthermore, the parallel generation of action sequences substantially improves inference efficiency compared to sequential autoregressive decoding.
To further enhance the model's comprehension of expert demonstration videos, we introduce a temporal-spatial masking strategy that stochastically masks video tokens across both temporal and spatial dimensions. This approach reduces computational complexity during training while concurrently establishing a more challenging learning objective that necessitates action prediction from partially observed expert demonstrations, thereby fostering holistic video understanding.

(\uppercase\expandafter{\romannumeral4}) 
\textbf{Dataset: Scarcity of expert-agent pair data.}
Training the generalizable \nickname model necessitates rich and diverse expert-agent pair data, which remain scarce in the robot learning domain, typically human-robot pair data. To address this data scarcity challenge, we develop a video-driven generation pipeline that synthesizes human-robot pairs from readily accessible human demonstration videos. The pipeline first grounds interaction information within human videos, and employs 3D Gaussian splatting to render realistic 4D scenes depicting an agent robot executing the demonstrated tasks, then produces observation-action data. Human-robot training pairs are constructed by pairing human videos with the generated robot demonstrations for the same task.
We collect 7,421 human videos covering over 100 distinct manipulation tasks, and construct the \dataname dataset containing 89,736 human-robot paired training samples through this pipeline.
Additionally, we utilize the open source datasets and construct pairs between data with similar tasks, yielding 803,175 paired samples. In total, we culminate 892,911 expert-agent samples for training \nickname.

Building upon the aforementioned insights, we introduce a novel VLA paradigm for robotic learning, termed \nickname, \textcolor{\mycolor}{which empowers robots to learn novel tasks from a single expert demonstration video at test-time.}
The \nickname framework comprises two key stages: 
(\uppercase\expandafter{\romannumeral1}) Latent Action Learning with Action-Centric Cycle-Consistency (A3C). We develop a latent action tokenizer that extracts latent action representations from observation sequences, incorporating action-centric cycle-consistency constraints to establish a unified latent action space spanning both expert videos and agent demonstrations. 
(\uppercase\expandafter{\romannumeral2}) \textcolor{\mycolor}{\nickname Training for One-Shot Task Learning.} The \nickname model is trained to predict latent action sequences for both expert videos and agent observations through parallel decoding, conditioned on expert demonstration videos processed via a temporal-spatial masking strategy, in conjunction with language instructions and agent observations.
To enhance the performance of \nickname, we develop a video-driven expert-agent paired data generation pipeline capable of generating paired samples from easily accessible human videos.
Additionally, we leverage the publicly available datasets to construct pairs between data with similar tasks. In total, we culminate 892,911 expert-agent paired samples for training our \nickname.
Experimental results demonstrate that our approach achieves more than 30\% improvement on unseen tasks in the LIBERO benchmark and exhibits over 35\% improvement using videos from different embodiments. Our method effectively distills knowledge from human videos, consistently yielding improvements exceeding 38\% on real-world unseen tasks.

Our contributions can be summarized as follows: 
(\uppercase\expandafter{\romannumeral1}) 
We propose a novel VLA paradigm, \nickname, which is able to effectively distill fine-grained manipulation knowledge from expert behavior and transfer it seamlessly to the agent. This paradigm enables policy models to acquire novel manipulation skills at test time from a single demonstration without necessitating further training or fine-tuning.
(\uppercase\expandafter{\romannumeral2}) We introduce a latent action learning framework incorporating cycle-consistency constraints to establish a unified latent action space that encompasses both expert demonstration videos and robot demonstrations. Furthermore, we employ a parallel decoding mechanism to mitigate the shortcut learning issue and enhance computational efficiency during inference.
(\uppercase\expandafter{\romannumeral3}) 
We present a scalable expert-agent pair data generation pipeline that synthesizes trajectory pairs from easily available video sources and integrates curated examples from publicly available datasets. Leveraging this pipeline, we construct a large-scale dataset containing 892,911 expert-agent paired trajectories across a diverse range of manipulation tasks.
(\uppercase\expandafter{\romannumeral4}) Experiments demonstrate that \nickname achieves over 30\% improvement on unseen tasks in the LIBERO benchmark and 35\% improvement when leveraging videos from different embodiments. Furthermore, our approach effectively distills knowledge from human videos, demonstrating improvements exceeding 38\% on real-world unseen tasks.

\section{Related work}
\label{sec:Related work}

\subsection{Vision Language Action Models}
Building upon the success of pretrained vision foundation models, large language models (LLMs), and vision-language models (VLMs), Vision-Language-Action models (VLAs) ~\cite{RT-2, openvla, qu2025spatialvla, pi_0, bjorck2025gr00t, pertsch2025fast, bu2025univla} have emerged as a promising approach for processing multimodal inputs to generate robotic actions. These methods leverage pretrained vision-language models and adapt them for robotic manipulation, effectively transferring semantic knowledge acquired from web-scale pretraining to robotics applications.
A pioneering work, RT-2~\cite{RT-2}, introduced a discretization strategy that uniformly quantizes continuous action values into 256 bins per dimension. This approach enables the co-training of web-scale language models on robot trajectory data, facilitating the transfer of semantic understanding to manipulation tasks. Building on this foundation, OpenVLA~\cite{openvla} adopts a similar action discretization approach while training vision-language models on the large-scale Open X-Embodiment (OXE) dataset~\cite{o2024open}, which aggregates robotic data from 22 distinct embodiments spanning 21 institutions. 
An alternative direction \cite{intelligence2025pi_, pi_0, bu2025agibot} employs action experts to generate continuous signals. $\pi_0$~\cite{pi_0} adapts the PaliGemma VLM by integrating a specialized action expert module that generates continuous actions via flow matching, enabling precise and fluent manipulation skills.
While Vision-Language-Action (VLA) models have demonstrated promising capabilities, existing approaches still struggle to generalize to tasks beyond their training distributions.
This constraint stands in contrast to human capabilities, where individuals learn novel manipulation knowledge by seamlessly visually imitating others. 
Motivated by this observation, we take a step forward by endowing VLAs with the ability to learn skills from a single expert demonstration video at test time.

\begin{figure*}[t]
     \begin{center}
    \includegraphics[width=1.0\textwidth]{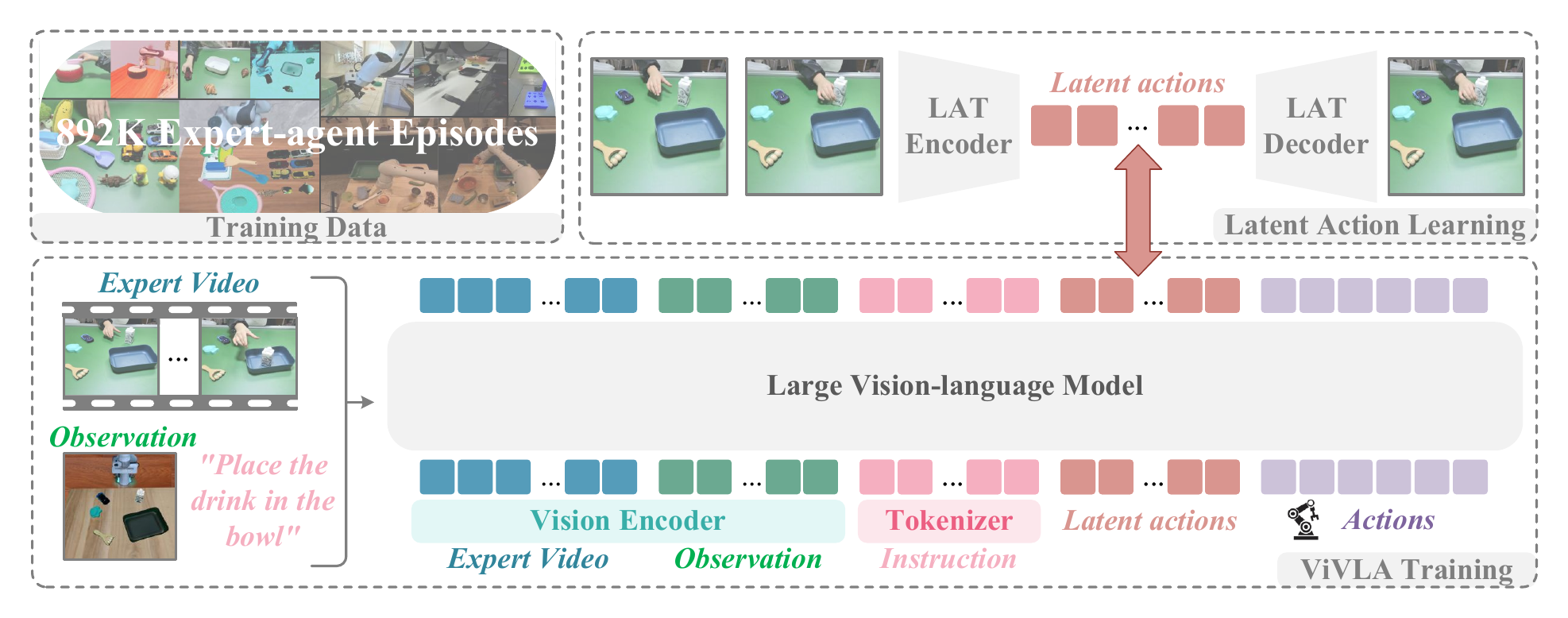}
    \end{center}

       \caption{{Overview of our \nickname. (\uppercase\expandafter{\romannumeral1}) The latent action tokenizer (LAT) learns quantized latent actions from observation sequences, obtaining latent actions for both expert videos and agent demonstrations. (\uppercase\expandafter{\romannumeral2}) The \nickname model is trained to predict the learned latent action sequences and subsequent robot actions, enabling the robot to acquire novel manipulation skills from only a single expert demonstration video at test time.}}
    \label{Fig::Overview}
\end{figure*}

\subsection{One-Shot Imitation Learning}
To enable agents to rapidly adapt to novel tasks from minimal demonstrations while incorporating human instructions at test time, Duan et al. \cite{duan2017one} first formulated the one-shot imitation learning (OSIL) problem. They proposed a soft-attention-based learning framework for block stacking tasks, where an agent processes both a single successful demonstration and the current observation to predict actions. Building upon this foundation, subsequent research has explored diverse demonstration formats, including cross-embodiment demonstrations~\cite{finn2017model, yu2018one}, demonstrations from varying viewpoints~\cite{sharma2019third}, and demonstrations paired with natural language instructions~\cite{jang2022bc, ahn2022can}.
To enhance task adaptation from these demonstrations, researchers have pursued multiple solution paradigms. These include task embedding conditioning~\cite{james2018task, bonardi2020learning}, meta-learning frameworks~\cite{finn2017model,yu2018one}, sub-goal prediction mechanisms~\cite{pathak2018zero, sharma2019third}, expressive transformer architectures~\cite{dasari2021transformers, mandi2022towards}, and contrastive learning of visual representations~\cite{mandi2022towards}. Finn et al. \cite{finn2017one} extended model-agnostic meta-learning (MAML) to visual imitation learning, enabling robots to rapidly adapt to new manipulation tasks from single demonstrations. T-OSVI \cite{dasari2021transformers} combined transformer-based attention mechanisms with self-supervised inverse dynamics losses to learn manipulation skills from single demonstration videos. AWDA \cite{chang2023one} improved generalization through a combination of attributed waypoint generation, demonstration augmentation, and image mixup techniques. More recently, OSVI-WM \cite{goswami2025osvi} proposed learning world models from expert demonstration videos by encoding visual observations into a shared latent space and predicting future latent trajectories, which are subsequently decoded into physical waypoints for robot execution.

\subsection{Learning from Cross-embodiment Data}
A promising goal of one-shot visual imitation learning is to enable robots to acquire skills directly from cross-embodiment demonstration videos, particularly human demonstrations, which offer the advantage of being readily accessible and providing an intuitive means for users to guide robotic behavior. 
To address the challenges of cross-embodiment learning, including substantial variations in camera perspectives, proprioceptive inputs, and action spaces across different embodiments, numerous approaches have been proposed.
Early efforts ~\cite{yang2024pushing} attempt to bridge these gaps through manual alignment of action spaces. Recent transformer-based methods~\cite{team2024octo, doshi2024crossformer} have been developed to accommodate variable observations and actions more effectively. CrossFormer~\cite{doshi2024crossformer} demonstrates the capability to co-train across four distinct action spaces without imposing constraints on observation spaces or requiring explicit action-space alignment.
Flow representations, which capture future trajectories of query points in images or point clouds, have been explored for cross-embodiment
learning~\cite{wen2023any,yuan2024general,gao2024flip,xu2024flow}. ATM~\cite{wen2023any} leverages video demonstrations for pre-training trajectory generation models, utilizing annotations from tracking models~\cite{karaev2024cotracker3, karaev2025cotracker, doersch2023tapir}. 
More recently, latent action-based methods~\cite{ye2024latent, bu2025univla, chen2024moto, bruce2024genie, chen2024igor} are proposed to learn a discrete codebook, which exhibits greater suitability for the autoregressive training paradigm inherent to VLAs. LAPA~\cite{ye2024latent} introduces an unsupervised learning framework for quantized latent actions between successive video frames, training models to reconstruct subsequent frames using latent actions and current frames. UniVLA~\cite{bu2025univla} proposes task-centric latent actions that incorporate instructions to decompose transition dynamics into task-irrelevant and task-relevant components.
Although these approaches have demonstrated promising performance, they remain limited in learning semantically consistent latent action representations. Moreover, the latent action spaces of different embodiments are typically fragmented, constraining manipulation knowledge transfer across embodiments.

\begin{figure*}[t]
 \setlength{\abovecaptionskip}{-0.10cm}
    \begin{center}
    \includegraphics[width=0.98\textwidth]{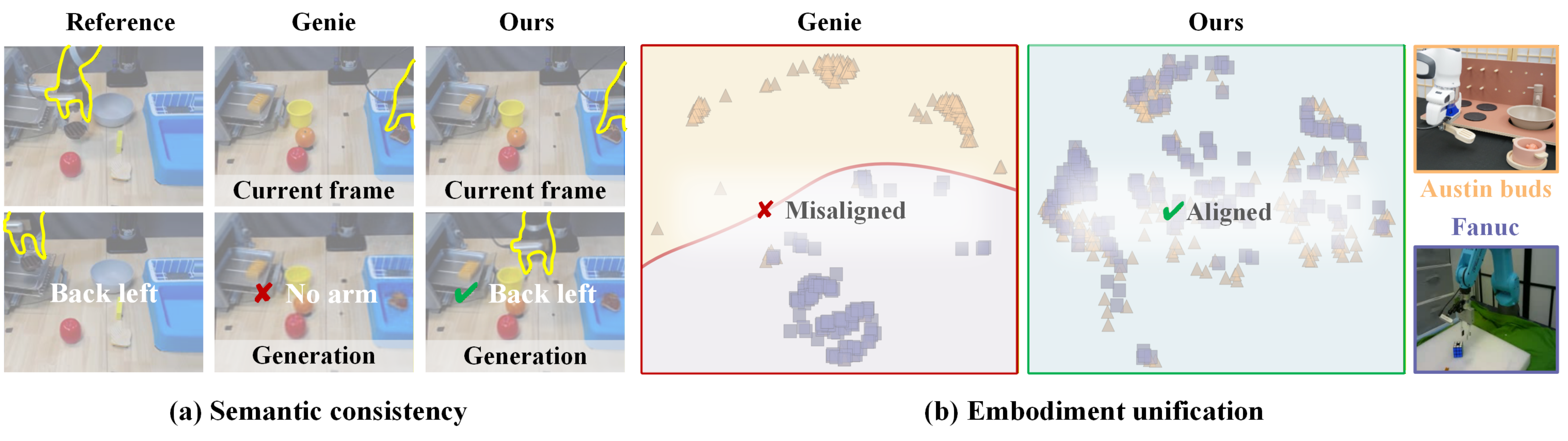}
    \end{center}
       \caption{{Motivation for our action-centric cycle consistency. (a) We apply latent actions encoded from reference video frames to the current frame. Existing methods, such as Genie \cite{bruce2024genie}, generate frames with divergent motion, revealing limited semantic consistency. (b) We visualize latent action spaces across embodiments, revealing limited cross-embodiment alignment in existing methods. Our method addresses these limitations and constructs a unified latent action space.}  }
    \label{Fig::Cycle_Fig1}
\end{figure*}

\subsection{Data Augmentation for Policy Learning}

In scenarios with limited training data, data augmentation has emerged as a promising approach to generate diverse training samples and enhance policy robustness. Prior research has explored diverse augmentation techniques to improve the resilience of visuomotor policies~\cite{laskin2020reinforcementlearningaugmenteddata, kostrikov2021imageaugmentationneedregularizing, mandlekar2021matterslearningofflinehuman, mandlekar2023mimicgendatagenerationscalable, fan2021secantselfexpertcloningzeroshot, hansen2021generalizationreinforcementlearningsoft, hansen2021stabilizingdeepqlearningconvnets}. A representative method is MimicGen~\cite{mandlekar2023mimicgen}, which presents an automated system for synthesizing large-scale robotic demonstration datasets from a limited number of human demonstrations. This is achieved by decomposing demonstrations into object-centric segments, spatially transforming these segments and subsequently stitching them together to generate novel trajectories. However, these approaches are predominantly evaluated in simulated environments, requiring task-specific environment construction and facing challenges related to the sim-to-real gap.
To facilitate the deployment of learned policies in real-world settings, recent work has explored augmenting scene appearance through image inpainting models~\cite{yu2023scalingrobotlearningsemantically, chen2024semanticallycontrollableaugmentationsgeneralizable, chen2023genaug, mandi2023cactiframeworkscalablemultitask}. For instance, Mirage~\cite{chen2024mirage} masks the target robot in images and inpaints the source robot at corresponding poses using URDFs and rendering techniques, enabling the transfer across different robot arms and grippers without requiring target robot training data. Similarly, VISTA~\cite{tian2024view} generates augmented task demonstrations from diverse camera viewpoints through single-image novel view synthesis models, aiming to learn view-invariant policies~\cite{ameperosa2024rocoda}. Furthermore, Rovi-aug~\cite{chen2024roviaugrobotviewpointaugmentation} develops a cross-embodiment pipeline by inpainting robot embodiments into image observations. Nevertheless, these studies primarily perform augmentation on 2D images, which inherently lack spatial information.
To address this limitation, RoboSplat~\cite{yang2025novel} reconstructs scenes using 3D Gaussian Splatting and edits the 3D representation for data augmentation.
In this work, we propose a video-driven expert-agent pair data generation pipeline. Our pipeline takes human videos as input and employs Gaussian splatting to render robot execution processes. Expert-agent pairs are constructed by pairing human videos with their generated robot demonstrations for the same task.

\begin{figure*}[t]
     \begin{center}
    \includegraphics[width=0.93\textwidth]{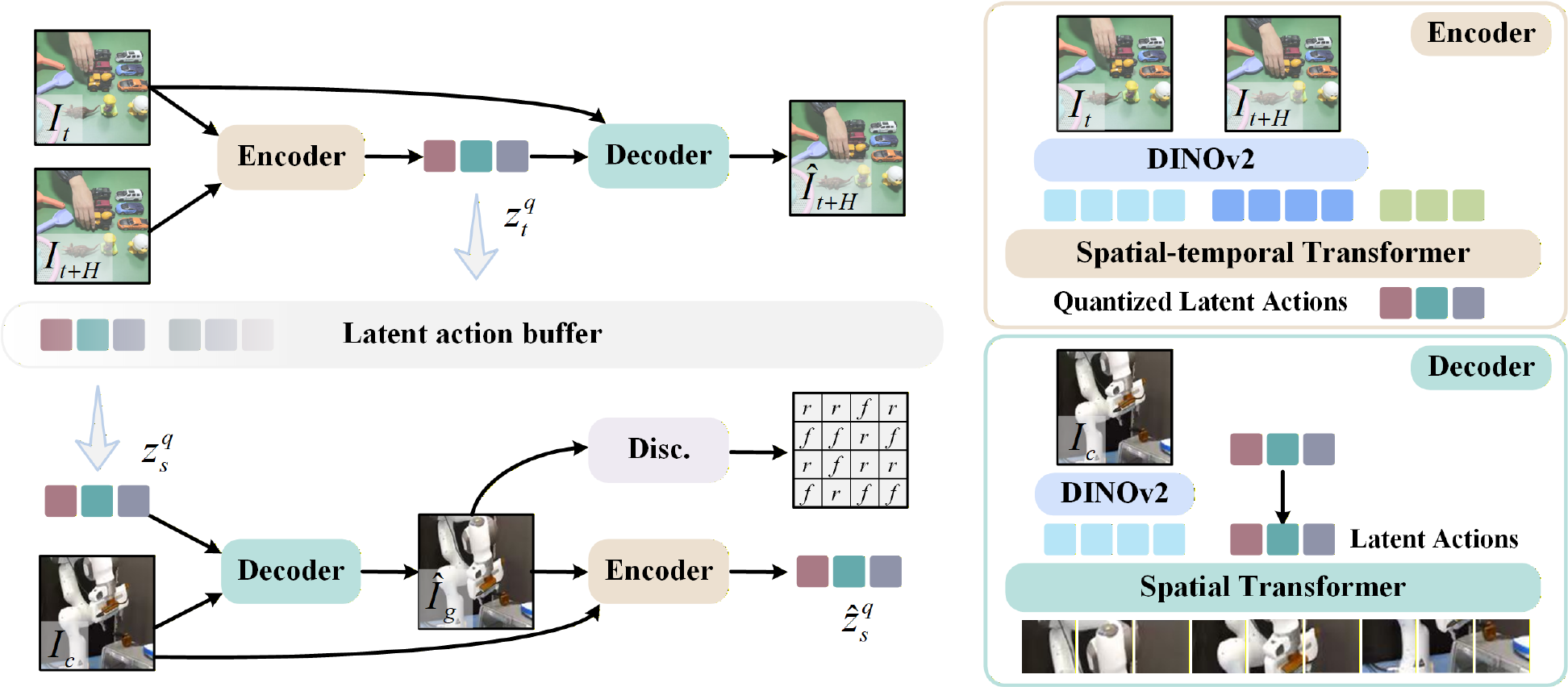}
    \end{center}

       \caption{{Illustration of our latent action framework with action-centric cycle consistency. Our approach learns latent action representations from observation frames, while simultaneously introducing action-centric cycle-consistency constraints to establish a unified latent action space.}  }
    \label{Fig::A3C_Overview}
\end{figure*}

\section{\nickname}
 Let $\mathcal{M}=\{\mathcal{M}_1, ..., \mathcal{M}_{|\mathcal{M}|}\}$ denote a set of tasks, partitioned into disjoint training and testing sets ($\mathcal{M}_{train}$ and $\mathcal{M}_{test}$). Each training task comprises a set of expert trajectories $\mathcal{T}^e=\{(v_{i}, \ell_{i})\}_{i=1}^{N_e}$, alongside an agent demonstration corpus annotated with actions $\mathcal{T}^a=\{(\tau^a_{i}, \ell_{i})\}_{i=1}^{N_a}$, each episode $i$ is paired with its corresponding language instruction $\ell_{i}$. The expert demonstration trajectory comprises expert video frames $\{v_{i,t}\}_{t=1}^{T_e}$, whereas agent demonstration $\tau^a_{i}=\{o_{i, t}, a_{i, t}\}_{t=1}^{T_a}$ encompasses both observation data $o$ and their associated action commands $a$. Within each task, all expert and agent trajectories correspond to variations  (e.g., different object configurations) of the same high-level task. The model, trained on $\mathcal{M}_{train}$, is evaluated on $\mathcal{M}_{test}$, which contains only expert videos.

Fig. \ref{Fig::Overview} presents the overall architecture of our \nickname. 
Our recipe for \nickname consists of two key stages: (\uppercase\expandafter{\romannumeral1}) Latent action learning with cycle-consistency. Our approach learns latent action representations from observation sequences, while introducing action-centric cycle-consistency constraints to establish a unified latent action space. (\uppercase\expandafter{\romannumeral2}) \nickname training for one-shot task learning.
The \nickname model is trained to predict action sequences for both expert videos and agent observations via parallel decoding, conditioned on expert demonstration videos processed with a temporal-spatial masking strategy, along with language instructions and agent observations.

\subsection{Latent Action Learning with Cycle Consistency} \label{Sec:: Graph_construction}
We learn latent actions from both agent demonstrations and expert videos, thereby providing latent action annotations for these two data sources.
Existing approaches learn latent action representations with a temporal window size $H$, achieved by modeling transition dynamics between successive observation frames $I_t$ and $I_{t+H}$, training the latent action tokenizer to reconstruct subsequent frames $I_{t+H}$ conditioned on both the learned latent actions and current frames $I_t$. 
However, the correspondence between current and future frames is strictly one-to-one, exclusively pairing $I_t$ with $I_{t+H}$. This permits future frame reconstruction with little understanding of transition dynamics, impeding the extraction of semantically consistent latent actions, as demonstrated in Fig. \ref{Fig::Cycle_Fig1}(a). 
Moreover, we visualize latent action spaces on distinct embodiments in Fig. \ref{Fig::Cycle_Fig1}(b) and observe fragmentation within these representational spaces, where distinct regions are assigned to individual embodiments, indicating limited cross-embodiment unification.

To overcome such limitations and establish a unified latent action space for \nickname training, we propose a latent action learning framework with action-centric cycle consistency and learns latent actions jointly on both expert videos and agent demonstration data. For brevity, we use $I$ to denote visual frames from both expert videos $v$ and agent observations $o$. As illustrated in Fig. \ref{Fig::A3C_Overview},  we extract latent actions $z^q_t$ from observation frames $\{I_t, I_{t+H}\}$ and maintain them in a latent action buffer $\mathcal{Z}$. 
We concurrently enforce action-centric cycle consistency, where latent actions $z^q_s$ sampled from this buffer are decoded with observation frame $I_c$ to generate future frames $\hat{I}_g$, and the tokenizer is trained to predict these sampled latent actions from both the observation frames $I_c$ and generated frames $\hat{I}_g$. As demonstrated in Fig. \ref{Fig::Cycle_Fig1}, our approach facilitates the construction of a unified latent action space that achieves both semantic consistency and cross-embodiment unification.

\textbf{Latent action tokenization}. 
Our method begins by learning latent actions from observation frames through an encoder-decoder architecture. The encoder $\mathcal{E}$ extracts image embeddings $f_t$ and $f_{t+H}$ from both current frames $I_t$ and future frames $I_{t+H}$ using DINOv2~\cite{oquab2023dinov2}. These extracted embeddings are concatenated with learnable latent action tokens. The combined representations are then processed by a spatial-temporal (ST) transformer, which consists of $L$ spatiotemporal blocks, each incorporating interleaved spatial and temporal self-attention layers. The ST-transformer models temporal transition dynamics between frames and aggregates the learned information into the latent action tokens.
The encoded latent action tokens $z^e_t$ are quantized into discrete representations $z^q_t$, where each latent action is represented using $l_{z}$ tokens selected from a codebook vocabulary of size $K$. These quantized action tokens are optimized using the VQ-VAE~\cite{van2017neural} objective.
The encoding procedure for the latent action $z^q_t$ is illustrated below:
\begin{equation}
\small
\setlength{\abovedisplayskip}{2pt}
\setlength{\belowdisplayskip}{2pt}
\begin{split}
z^e_t &= \text{ST-Transformer}([f_t, f_{t+H}, z]), \quad z^e_{t} \in \mathbb{R}^{l_z \times c_z} \\
z^q_t &= \text{VQ}(z^e_t), \quad z^q_{t} \in \mathbb{R}^{l_z \times c_z}.
\end{split}
\end{equation} 
The decoder $\mathcal{D}$, implemented as a spatial transformer containing spatial blocks with spatial attention layers, reconstructs the future frames $I_{t+H}$ by processing the learned latent actions $z^q_t$ alongside the current frames $I_t$. In summary, the latent action tokenization procedure is formulated as follows:
\begin{equation}
\small
\setlength{\abovedisplayskip}{2pt}
\setlength{\belowdisplayskip}{2pt}
\begin{split}
z^q_t &= \mathcal{E}(I_t, I_{t+H}), \quad z^q_{t} \in \mathbb{R}^{l_z \times c_z}, \\
\hat{I}_{t+H} &= \mathcal{D}(I_t, z^q_t), \quad \hat{I}_{t+H} \in \mathbb{R}^{w \times h \times c_o}.
\end{split}
\end{equation} 

\begin{figure*}[t]
     \begin{center}
    \includegraphics[width=1.0\textwidth]{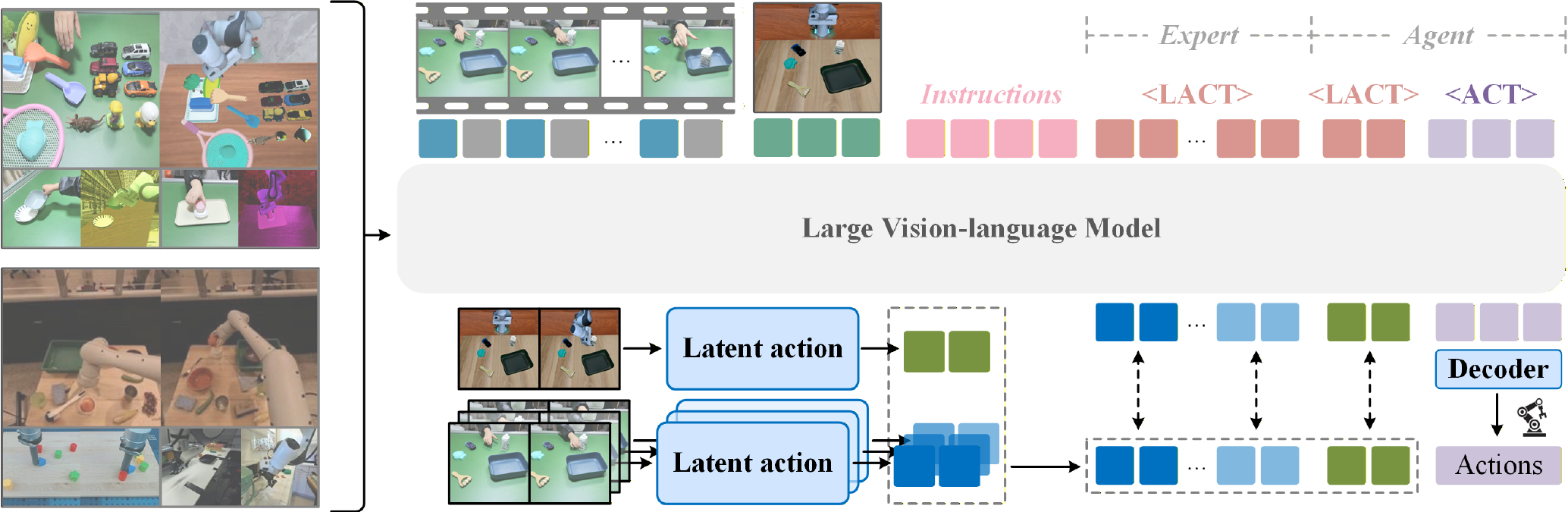}
    \end{center}

       \caption{{Illustration of our \nickname. Our approach jointly processes an expert demonstration video alongside the robot's visual observations to predict both the demonstrated action sequences and subsequent robot actions, facilitating the \nickname to distill fine-grained manipulation knowledge from expert behavior and transfer it seamlessly to the agent. }  }
    \label{Fig::VideoVLA_Overview}
\end{figure*}

\textbf{Action-centric cycle consistency}.
We introduce action-centric cycle consistency to regularize the learned latent actions, establishing a unified latent action space that achieves both semantic consistency and cross-embodiment unification.
Our method takes observation frames $I_c$ from the dataset and samples a latent action $z^q_s$ from the latent action buffer $\mathcal{Z}$, where the latent action buffer $\mathcal{Z}$ is constructed by collecting encoded latent actions $z^q_t$ over the previous $B$ batches. We then decode the observation frames $I_c$ with the sampled latent action $z^q_s$ to generate the corresponding subsequent frame $\hat{I}_g$. 
The cycle consistency is enforced by providing both the observation frames $I_c$ and generated frames $\hat{I}_g$ into the encoder $\mathcal{E}$, which is trained to recover the originally sampled latent action $z^q_s$:
\begin{equation} \label{Eq::Graph}
\setlength{\abovedisplayskip}{5pt}
\setlength{\belowdisplayskip}{5pt}
\begin{split}
\hat{I}_g &= \mathcal{D}(I_c, z^{q}_{s}), \quad z^q_s \sim \mathcal{Z},\\
\hat{z}^q_s &= \mathcal{E}(I_c, \hat{I}_g), \quad \hat{z}^q_s \in \mathbb{R}^{l_z \times c_z}.\\
\end{split}
\end{equation}
To enable gradient propagation, we utilize the latent action embeddings $\hat{z}^e_s$ before quantization and compute their distances to codebook vectors $e$ in the VQ-VAE as similarity measures. We then optimize consistency with the sampled latent actions $z_s^q$ through cross-entropy loss, using the codebook indices of sampled latent actions as supervision signals. The objective is formulated as follows:
\begin{equation} \label{Eq::Graph}
\setlength{\abovedisplayskip}{5pt}
\setlength{\belowdisplayskip}{5pt}
\begin{split}
\mathcal{L}_C = -\Sigma_{k=1}^{K} y_k log(\frac{exp(-d(\hat{z}^e_s, e_k)/\tau)}{\Sigma_{j=1}^{K} exp(-d(\hat{z}^e_s, e_j)/\tau)}).
\end{split}
\end{equation}
where $y_k$ is the one-hot target vector corresponding to the indices of sampled latent actions $z_s^q$, $\tau$ is the temperature parameter, $\hat{z}^e_s$ is the latent action embedding prior to quantization, $\{e_k\}_{k=1}^K$ are the codebook vectors, and $d(\cdot, \cdot)$ denotes the distance metric. Unlike previous methods that are constrained by fixed pairing patterns between current and future frames, limiting the effectiveness of latent action acquisition, our method establishes a challenging self-supervised learning objective. This formulation compels the tokenizer to extract semantically consistent latent actions. Furthermore, our method enables cross-embodiment learning to facilitate the unification of the latent action space across embodiment,s. Specifically, we sample latent actions $z_s^q$ encoded from embodiment ${E}_i$ and apply them to observation frames $I_c$ from embodiment ${E}_j$, generating frames $\hat{I}_g$ and predicting latent actions $\hat{z}_s^q$ for ${E}_j$. Cycle consistency enforces $\hat{z}_s^q \approx z_s^q$ to compel the tokenizer to learn the unified latent actions across embodiments.

\textbf{Discriminator}. 
The distributional discrepancy between decoder-generated images and dataset images degrades encoder performance.  Furthermore, the decoder may leak latent action information directly to the encoder through the generated frame, resulting in correspondences that appear cycle-consistent yet remain fundamentally erroneous.
To address this issue, we introduce a local-global discriminator that aligns the distribution of generated frames with dataset images across both local details and global style. 
This regularization simultaneously prevents information leakage since embedding latent actions into generated frames induces distributional deviations that the discriminator penalizes.
The discriminator $\Psi$ processes input image frames, including both observation frames from datasets and generated frames, through the spatial transformer to extract corresponding patch features $h_l$, which are subsequently processed by an MLP to produce patch logits $\sigma_l$. Additionally, we apply 2D convolution and global pooling operations to the patch features to derive global features $h_g$, from which global logits $\sigma_g$ are obtained, formally expressed as:
\begin{equation} \label{Eq::Graph}
\setlength{\abovedisplayskip}{5pt}
\setlength{\belowdisplayskip}{5pt}
\begin{split}
\mathcal{F}_l &= \text{Spatial-Transformer}(I), \quad \sigma_l = \text{MLP}(\mathcal{F}_l), \\
\mathcal{F}_g &= \text{GlobalPool}(\text{Conv2D}(\mathcal{F}_l)), \quad \sigma_g = \text{MLP}(\mathcal{F}_g). \\
\end{split}
\end{equation}
Based on these local and global logits, we define the adversarial losses $\mathcal{L}_{GAN}$ for both the decoder $\mathcal{D}$ and discriminator $\Psi$, which are formulated as follows:
\begin{equation} \label{Eq::Graph}
\setlength{\abovedisplayskip}{5pt}
\setlength{\belowdisplayskip}{5pt}
\begin{split}
\mathcal{L}^{\Psi}_{GAN} &= -log(\Psi(o)) - (1-log(\Psi(\mathcal{D}(o, z)))), \\
\mathcal{L}^{\mathcal{D}}_{GAN} &= 1-log(\Psi(\mathcal{D}(o, z))). \\
\end{split}
\end{equation}
These loss formulations are applied across both local and global levels. The objective of the discriminator is to maximize the probability of correctly distinguishing between dataset samples and decoder-generated samples. The decoder is trained to maximize the probability that the discriminator classifies its generated samples as originating from the dataset distribution.

\subsection{\nickname Training for One-Shot Task Learning}

The overview of our \nickname are illustrated in Fig. \ref{Fig::VideoVLA_Overview}. Our framework builds upon the Qwen2.5-VL \cite{bai2025qwen2} vision-language model, which has demonstrated promising performance in visual recognition and semantic-level video understanding. The architecture consists of a Vision Transformer (ViT) with window attention for efficient processing at native resolutions, an MLP-based vision-language merger that compresses visual features, and the Qwen2.5 large language model that excels at multi-modal understanding. 
We initialize the weights with pre-trained parameters and train the model on the expert-agent pair data to enhance the action understanding and skill learning capabilities. 
The policy model receives expert demonstration videos with temporal-spatial masking, along with robot agent observations, and language instructions, aiming to predict action sequences demonstrated in expert video sequences as well as the subsequent actions performed by agents. 

\textbf{Temporal spatial masking strategy}.
Video data constitute natural signals characterized by substantial temporal and spatial redundancy. The incorporation of video data generates extensive token sequences that impose considerable computational burdens during training.
To mitigate these challenges, we propose a temporal-spatial masking strategy that masks video data across both temporal and spatial dimensions.
We apply temporal masking using the same temporal window size as Qwen2.5VL for input video sequences, while preserving the absolute time encoding corresponding to original timesteps to maintain the temporal information. The retained video frames are subsequently processed through the vision encoder, generating a set of patch-wise token representations. We then apply spatial masking to the resulting token representations, forwarding only unmasked tokens to the language model components. This temporal-spatial masking approach substantially reduces video information redundancy while compelling VLMs to perform action prediction on partially observed video sequences, thereby enhancing their capacity for comprehensive video understanding. Following this procedure, the processed video tokens, together with agent observation tokens and language instruction tokens, are encoded into latent representations $\{{h}_{v}, {h}_{o}, {h}_{\ell}\}$ for subsequent action prediction.

\textbf{Parallel decoding}. 
The encoded tokens $\{{h}_{v}, {h}_{o}, {h}_{ \ell}\}$ are fed into the language model components $\text{LM}$, which are trained to predict action sequences. To mitigate the shortcut learning problem and reduce inference latency, the language model employs parallel decoding to predict action sequences, encompassing both latent actions and robot actions. 
Specifically, for latent action prediction, we extend the vocabulary with latent action query tokens for latent action prediction, denoted as $\texttt{LACT}$. The model receives these latent action query tokens as inputs and simultaneously decodes them in a single forward pass. To enable the model to adaptively determine the initiation of parallel decoding and the number of tokens to be decoded concurrently, we introduce START tokens to the vocabulary: $\{\texttt{START\_LACT\_n} \mid n \in \{1,2,\ldots,\texttt{MAX\_N}\}\}$, where $\texttt{MAX\_N}$ indicates the maximum number of latent action query tokens. When the model generates START tokens, the corresponding latent action query tokens are appended to the subsequent input according to their designated parallel decoding quantity, generating all action tokens concurrently in a single forward pass. 
The parallel decoding process is represented as:
\begin{equation}
\begin{split}
    \small
    s_t &= \text{LM}({h}_{\leq t}), {h}_{\leq t} = [{h}_{v}, {h}_{o}, {h}_{ \ell}, {h}_{p}] \\
    \{\hat{z}_{t+i}\}_{i=0}^{n-1} &= \text{LM}([{h}_{\leq t}, s_t, \texttt{LACT}^n]).
\end{split}
\end{equation}
where ${h}_{p}$ represents previously generated tokens, $s_t \in \{\texttt{START\_LACT\_n} \mid n \in \{1,2,\ldots,\texttt{MAX\_N}\}\}$, $\texttt{LACT}^n$ indicates $n$ $\texttt{LACT}$ tokens, $\hat{z}$ denotes the predicted latent action tokens.
Similarly, robot action prediction follows the same parallel decoding mechanism, with the introduction of corresponding START tokens $\{\texttt{START\_ACT\_n} \mid n \in \{1,2,\ldots,\texttt{MAX\_N}\}\}$ and action query tokens $\texttt{ACT}$ to enable parallel prediction of action sequences.
In contrast to the autoregressive modeling strategy, which has access to all preceding ground truth action tokens during training and necessitates sequential token-by-token processing, inducing the shortcut learning issue and increasing inference latency.
The parallel decoding approach effectively prevents information leakage and encourages action prediction based on the understanding of the expert videos and agent observations. Furthermore, concurrent generation of all action sequences substantially enhances inference efficiency.

\textbf{\nickname training}. 
The training prediction objectives comprise the latent action sequences $\{z^q_{v,nH}\}_{k=0}^{N_v-1}$ from expert video sequences, along with the subsequent latent actions $z^q_{o,t}$ and robot actions of agent robots, where $N_v = \lfloor T/H \rfloor$, $T$ denotes the video length, and $H$ indicates the temporal window size of the latent action encoding.
The latent actions $z^q_{v,t}$ and $z^q_{o,t}$ corresponding to expert video frame $v_{t}$ and agent observation frame $o_{t}$ are encoded using the pre-trained latent action tokenizer, following the procedure outlined below:
\begin{equation}
\begin{split}
    \small
    z^q_{v,t} &= \mathcal{E}(v_{t}, v_{t+H}), \\
    z^q_{o,t} &= \mathcal{E}(o_{t}, o_{t+H}).
\end{split}
\end{equation}
These latent actions are represented using $l_{z}$ tokens selected from a codebook vocabulary of size $K$, which naturally aligns with the discrete prediction paradigm employed by VLMs. We extend the vocabulary by incorporating $K$ specialized tokens: 
$\{\texttt{{LACT\_1}, {LACT\_2}, {LACT\_3},..., {LACT\_K}\}}$. 
Each latent action is mapped to this extended vocabulary based on its corresponding index within the latent action codebook. The optimization objective centers on minimizing the sum of negative log-probabilities for subsequent latent actions:
\begin{equation}
\small
    \mathcal{L}_z = \mathbb{E}_{z} \left[ - \sum_{i=1}^{N_z} \log \ P(\hat{z}^q_{i} = z^q_{i}) \right],
\end{equation}
where $N_z$ represents the total length of latent action tokens.
The action query tokens ${\texttt{ACT}}$ are appended following the latent action query tokens for the robot action prediction, and an action decoder is integrated to transform the predicted action tokens ${\texttt{ACT}}$ into continuous robot action values. The action decoder receives action embeddings from the final layer of the $\text{LM}$ as input. It then aggregates this information through an attention mechanism and pools the representations into a unified embedding, which is subsequently mapped to robot actions via an MLP. The complete architecture is trained end-to-end by jointly optimizing the latent action prediction loss and the L1 loss between ground truth and predicted robot actions.

\textcolor{\mycolor}{To facilitate cross-modal information exchange between video and image representations, we incorporate a temporal localization task. Specifically, the agent's observation images are inserted into the expert demonstration video sequence, and the model is trained to identify their temporal positions within the video, thereby enhancing feature exchange across modalities.
Additionally, we randomly exclude the expert demonstration video during training, requiring the model to predict latent actions and robot actions based solely on robot observations and language instructions. This training strategy enables our policy to remain robust for seen tasks without access to expert demonstrations.}

\textbf{\nickname post-training}. 
To effectively transfer the knowledge acquired during pre-training to the target robotic platform, we conduct post-training of our \nickname model on the target robot. During this post-training phase, the action decoder undergoes full parameter fine-tuning to adapt to the specific action space and control requirements of the target robot, while the pre-trained VLM backbone is fine-tuned using Low-Rank Adaptation (LoRA) \cite{hu2022lora}. Such a design preserves the rich semantic representations and generalization capabilities learned during pre-training, while simultaneously enabling efficient adaptation to the target robotic domain with minimal computational overhead. The overall procedure follows the same framework established in the pre-training stage.

\begin{figure*}[t]
     \begin{center}
    \includegraphics[width=0.98\textwidth]{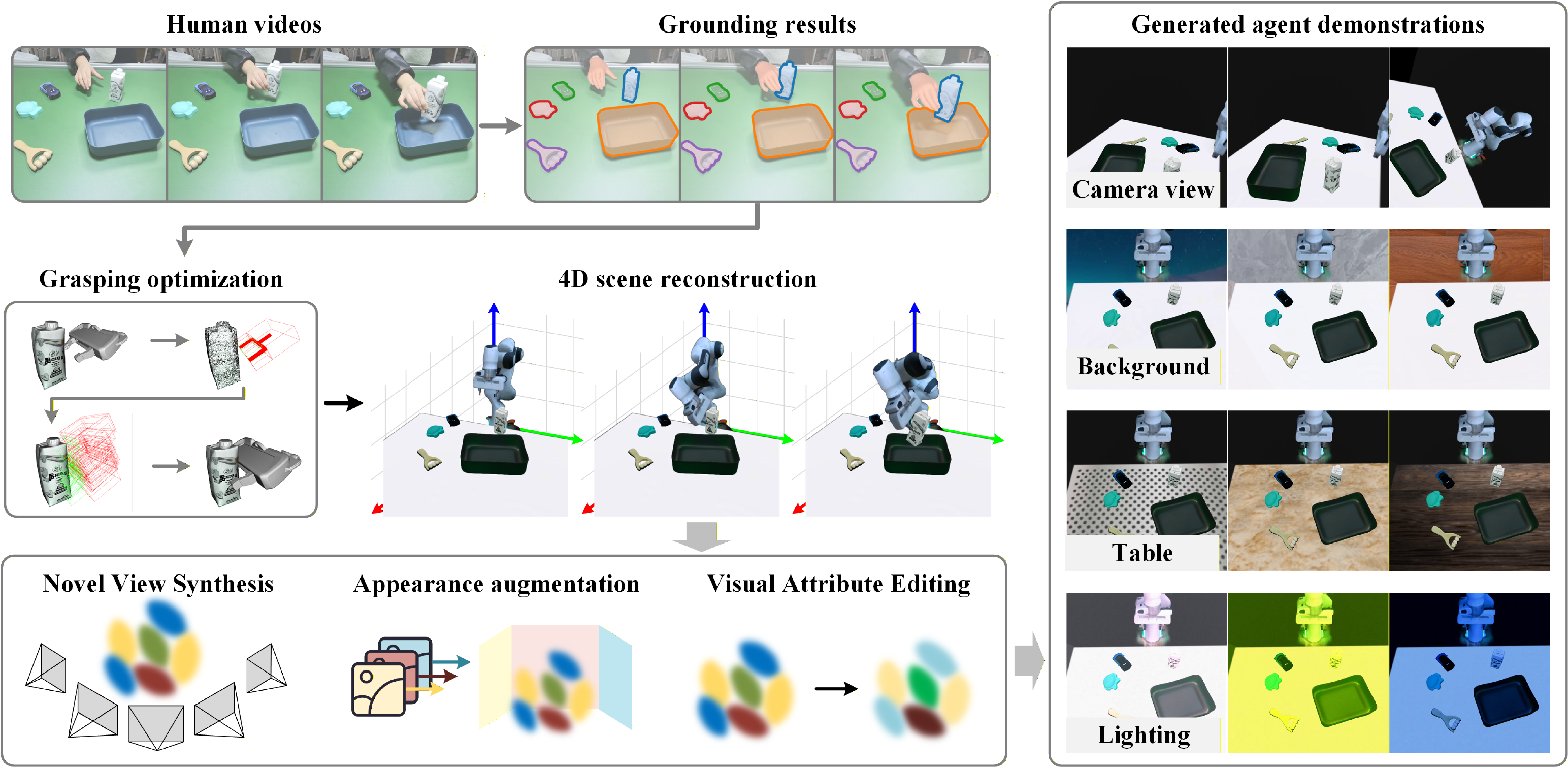}
    \end{center}
       \caption{{Illustration of our video-driven expert-agent data generation pipeline. This pipeline takes human videos as input, utilizes existing vision foundation models to obtain hand poses and object poses, and employs Gaussian splatting to reconstruct 4D scenes of different robots performing the tasks. Expert-agent training pairs are constructed by pairing human videos with their corresponding generated robot demonstrations for the same task.}  }
    \label{Fig::Data_overview}
\end{figure*}

\subsection{Video-driven Expert-agent Data Generation}\label{Sec:Data_Gen}
One of the keys to training generalizable robotic models lies in diverse and high-quality training data. To generate diverse and high-quality expert-agent pair data, we construct a video-driven expert-agent pair data generation pipeline, as shown in Fig. \ref{Fig::Data_overview}. This pipeline takes human videos as input, utilizes existing vision foundation models to obtain hand poses and object poses, and employs Gaussian splatting to reconstruct 4D scenes of the robot performing the tasks. Expert-agent pairs are constructed by pairing human videos with the generated robot demonstrations for the same task.

\begin{figure}[t!]
\setlength{\abovecaptionskip}{-0.13cm}
\begin{center}
\includegraphics[width=0.30\textwidth]{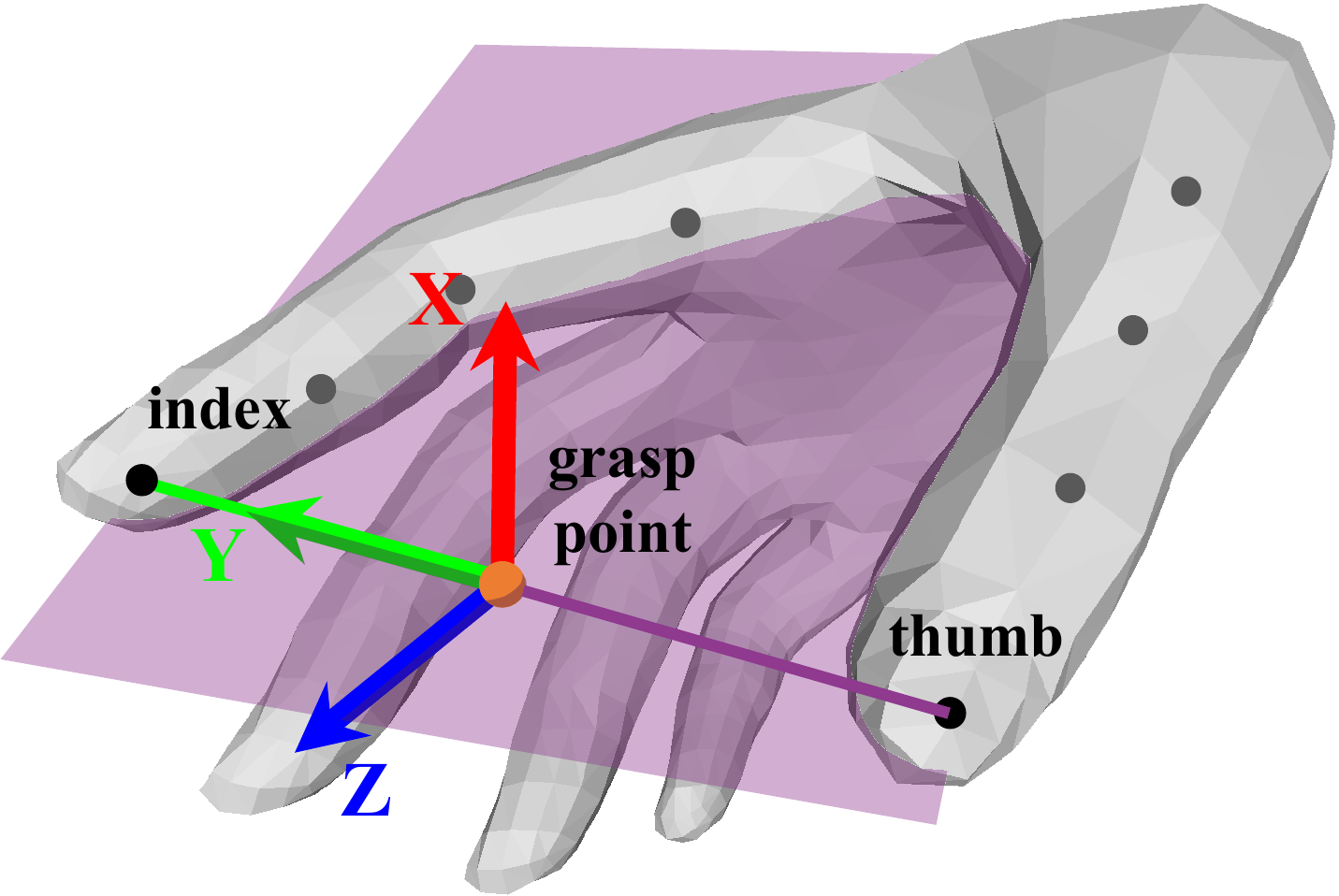}
\end{center}
\caption{Calculation of the 6D gripper pose from the estimated hand pose. The grasp point is computed as the midpoint between the thumb and index finger tips. The coordinate frame is defined with the $X$-axis normal to the plane spanned by all tracked points on both digits, the $Y$-axis pointing from the grasp point to the index finger tip, and the $Z$-axis obtained through $\vec{z}=\vec{x} \times \vec{y}$. }
\label{handpose}
\vspace{-8pt}
\end{figure}

\textbf{Interaction grounding}. 
We utilize vision foundation models to estimate hand and object poses from the provided human video $v$.
For hand tracking, we first apply HaMeR \citep{pavlakos2024reconstructing} to predict hand shape and pose parameters, while generating a hand mesh model. The Iterative Closest Point (ICP) \citep{besl1992method, rusinkiewicz2001efficient} is further implemented to align the hand mesh with the segmented hand point cloud, yielding precise hand pose trajectories $\bm{\bm{\xi}}_H =\{\bm{x}^{H^0}_{C},...,\bm{x}^{H^T}_{C}\}$ in the camera frame. Subsequently, these sequences are converted to robot end-effector trajectories $\bm{\xi}_E =\{\bm{x}^{E^0}_{C},...,\bm{x}^{E^T}_{C}\}$ \cite{chen2025fmimic}, as illustrated in Fig. \ref{handpose}.  
For object tracking, FoundationPose \citep{wen2023foundationpose} estimates temporal object poses $\bm{\xi}_\Omega =\{\bm{x}^{\Omega^0}_{C},...,\bm{x}^{\Omega^T}_{C}\}$ given the object mesh ${\bm \Omega}$, which is reconstructed from multi-view images via TRELLIS \cite{xiang2025structured}.

\textbf{Video parsing} 
We segment videos into individual clips $\{\bm{\tau}_i\}_{i=1}^{V} $, where each clip encapsulates a distinct subtask. 
This segmentation process is predicated on the identification and utilization of interaction markers, which denote the onset of contact and the termination of contact.
Specifically, the point clouds ${\mathcal{P}}$ of objects and hands are obtained through their respective pose estimation and mesh models.
We then compute inter-object distances and identify contact transitions as follows:
\begin{equation} 
\setlength{\abovedisplayskip}{2pt}
\setlength{\belowdisplayskip}{2pt}
\begin{split}
{d} &= {\rm dist}({\mathcal{P}}),  \quad {t}_b = \{t|{d}^{t-1}\! >\! \epsilon \wedge {d}^{t} < \epsilon\}, \\ {t}_e &= \{t|{d}^{t-1}\! <\! \epsilon \wedge {d}^{t} > \epsilon\},
\end{split}
\end{equation}
{where function ${\rm dist}$ calculates the distance between any two point clouds. ${t}_b$ and ${t}_e$ denote contact initiation and termination, respectively.}
We classify the clips into the grasping phase and the manipulation phase.
In the grasping phase, the objects remain stationary, and the agent executes a reach-and-grasp maneuver targeting the object.
In the manipulation phase, the agent manipulates the grasped object, performs a motion, and makes contact between objects.

\textbf{End-effector pose optimization}. 
We optimize the converted end-effector pose trajectories ${\xi}_E$ to further refine the contact relationship between the end-effector and the object.
During the grasping phase, we replicate the robot end-effector's trajectory using grounded trajectories and optimize the grasping pose ${x}^{E^{t_g}}_{C}$ at the moment of contact $t_g$, where $t_g$ denotes the termination of the grasping phase. Specifically, we sample $N$ candidate grasps $\{{x}^{E^{t_g}}_{C,i}\}_{i=1}^{N}$ within a 6D neighborhood around the initial grounded grasping pose ${x}^{E^{t_g}}_{C}$, and subsequently filter for feasible grasping configurations. A grasp is considered feasible if two conditions are satisfied: (\uppercase\expandafter{\romannumeral1}) the end-effector does not collide with the object, and (\uppercase\expandafter{\romannumeral2}) the object lies within the grasping region of the end-effector. For parallel-jaw grippers, the grasping region is typically defined as the region between the gripper jaws. Among feasible candidates, we compute a stability score for each grasp as the normalized perpendicular distance between the gripper plane and the object's center of gravity (COG) \cite{fang2023anygrasp}. The feasible grasping pose ${\hat{x}}^{E^{t_g}}_{C}$ with the highest stability score is then selected and expressed relative to the object's coordinate frame.
In the manipulation phase, we adopt the commonly used assumption that the relative pose between the end-effector and object remains invariant throughout the manipulation. Under this constraint, the end-effector trajectories are derived by composing the object's motion with the optimized grasping pose. The resulting optimized end-effector poses ${\hat{\xi}}_E$ enable our data generation pipeline to synthesize high-quality 4D robot–object manipulation scenes with improved physical consistency.

\textbf{High quality 4D scene reconstruction}.
The grounded object trajectories and optimized end-effector pose trajectories are used to reconstruct 4D robot–object manipulation scenes via 3D Gaussian splatting. During reconstruction, each object is represented using its grounded pose and corresponding Gaussian model $\mathcal{G}_{\Omega}$, while the robot is represented using joint positions and its Gaussian model $\mathcal{G}_{R}$. The robot joint positions are derived from the optimized end-effector trajectories ${\hat{\xi}}_E$ through motion planning, the object Gaussian models $\mathcal{G}_{\Omega}$ are converted from the object meshes, and the robot Gaussian models $\mathcal{G}_{R}$ are obtained from the corresponding URDF file or real-world reconstruction. The reconstructed 4D scenes exhibit high visual fidelity and align well with the trajectories observed in the human video $v$, as demonstrated in Fig. \ref{Fig::Data_overview}.

\begin{figure}[t!]
\setlength{\abovecaptionskip}{-0.13cm}
\begin{center}
\includegraphics[width=0.48\textwidth]{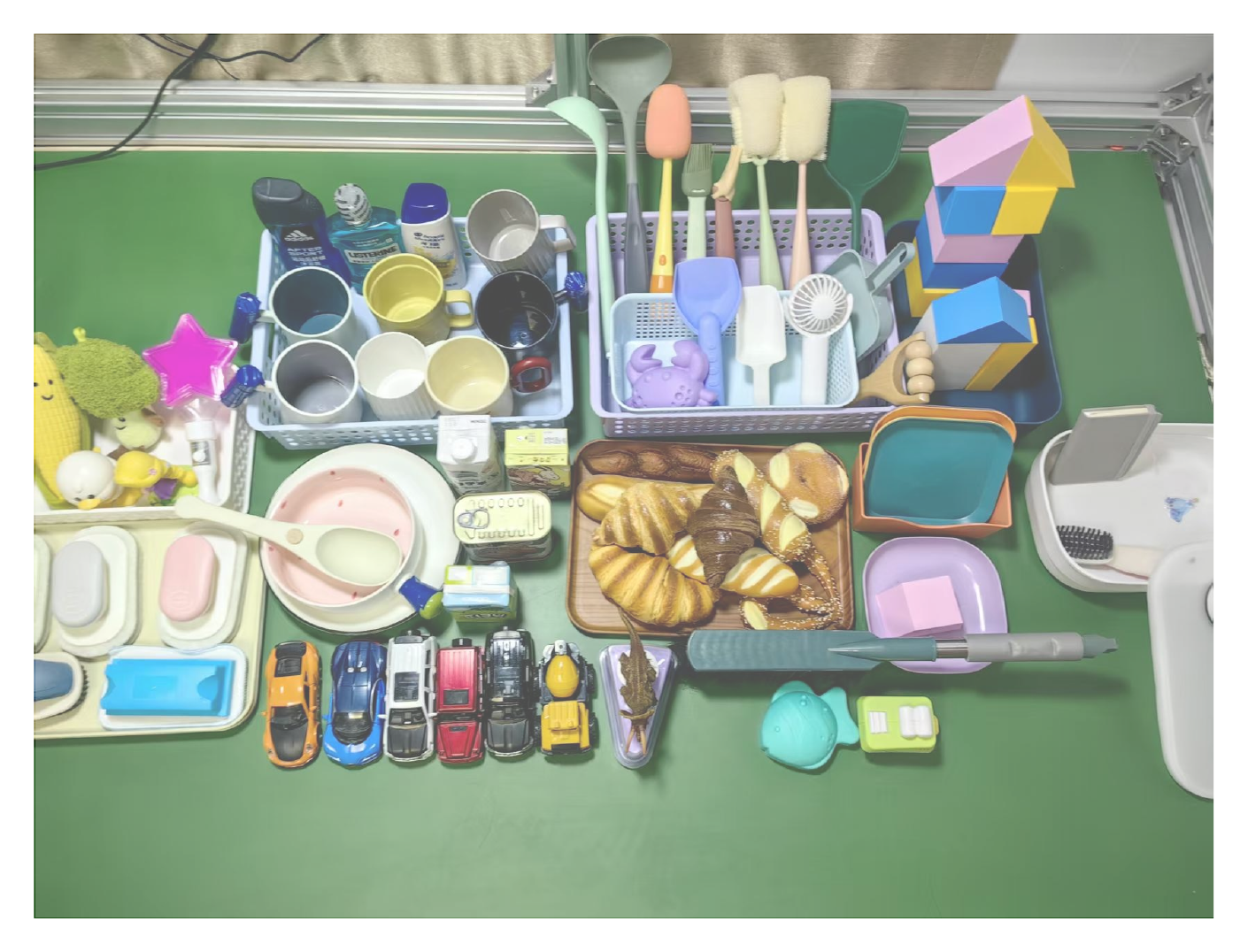}
\end{center}
\caption{The set of objects used for the data generation pipeline.}
\label{DATA_all}
\vspace{-8pt}
\end{figure}

\textbf{Diverse expert-agent data generation}. 
Leveraging the reconstructed 4D scene, we generate robot demonstrations through diverse demonstration augmentations \cite{yang2025novel}, which operate along three dimensions:
(\uppercase\expandafter{\romannumeral1}) {Camera view.} A key advantage of 3D Gaussian splatting lies in its capability for novel view synthesis. We exploit this property to generate demonstrations from multiple viewpoints by rendering the reconstructed 4D scene from diverse camera poses, thereby enriching the visual diversity of the training data.
(\uppercase\expandafter{\romannumeral2}) {Scene appearance.} We augment scene appearance by applying varied textures to the table and background 3D Gaussian planes surrounding the manipulation workspace. This enables the generation of diverse scene compositions that enhance the model's robustness to environmental variations.
(\uppercase\expandafter{\romannumeral3}) {Lighting condition.} We diversify lighting conditions by applying random scaling factors, offsets, and noise perturbations to the diffuse color attributes of each Gaussian in the reconstructed scene. 
We curate 7,421 human demonstration videos spanning over 100 distinct manipulation tasks, and construct the \dataname dataset containing 89,736 human-robot paired training samples through this pipeline. The objects used in the data collection process are visualized in Fig. \ref{DATA_all}.

\textbf{Dataset collection from publicly available dataset}. 
In addition to our generated dataset, we incorporate publicly available datasets to further enrich the training corpus by constructing expert-agent pairs from demonstrations of similar tasks. Task similarity is quantified by computing the cosine similarity between Sentence-BERT embeddings \cite{reimers-2019-sentence-bert} of task instructions, and we select pairs that exceed a predefined similarity threshold (set to 0.9 in our implementation).
Through this process, we collect 803,175 paired samples from publicly available datasets. In total, we culminate 892,911 expert-agent paired samples for training our \nickname.

\begin{table}[t]
    \centering
    \small
    \caption{Training data mixture.}
       \begin{tabular}{>{\raggedright\arraybackslash}m{5.50cm} | >{\raggedleft\arraybackslash}m{2.50cm}}
        \toprule
                Datasets & Num of Traj. \\
        \midrule
        Fractal~\cite{brohan2022rt1} & 87,212 \\
                Bridge~\cite{walke2023bridgedata} & 60,064\\
        Droid~\cite{khazatsky2024droid} & 49,933 \\
                                                                Language Table~\cite{lynch2023interactive} & 442,226 \\
                                                                                                        BC-Z~\cite{jang2022bc} & 43,264 \\
        FMB Dataset~\cite{luo2024fmb}  & 4,592 \\
                                                                Ego4D~\cite{khazatsky2024droid}  & 18,368\\
        EgoDex~\cite{hoque2025egodex} &  97,516\\
        \dataname & 89,736 \\
        \midrule
        Overall & 892,911\\
                \bottomrule
        \end{tabular}        \label{tab:data_mix}
\end{table}

\section{Implementation details} \label{Implementation}

Our training corpus consists of three primary data categories: robotic manipulation data, human video data, and synthetically generated human-robot pair data (our \dataname dataset). For robotic manipulation, we incorporate a selected subset from the Open X-Embodiment dataset~\cite{padalkar2023open}, specifically focusing on single-arm end-effector control tasks. The human video data comprises ego-centric recordings from the Ego4D dataset~\cite{grauman2022ego4d} and EgoDex~\cite{hoque2025egodex}, capturing first-person perspectives of diverse human activities. We leverage the demonstrations from robotic manipulation data and human video data by generating expert-agent pairs through the combination of trajectories with semantically similar task demonstrations. Our \dataname is synthesized through a video-driven expert-agent data generation pipeline, which we detail in Sec. \ref{Sec:Data_Gen}. A comprehensive breakdown of the dataset composition is provided in Table~\ref{tab:data_mix}.

For latent action learning with action-centric cycle consistency, our encoder leverages a pre-trained ViT-base DINOv2 model to extract visual embeddings from input images. These embeddings are subsequently processed through a spatial-temporal transformer consisting of 12 layers with 768-dimensional hidden states. The resulting latent action representations are discretized via vector quantization using a codebook containing 16 entries. Both the decoder and discriminator networks employ spatial transformers with identical architectural specifications, each comprising 12 layers and maintaining 768-dimensional representations throughout. To facilitate temporal consistency during training, we maintain a latent action buffer that accumulates encoded latent actions from the preceding 4 batches. The entire latent action tokenizer is optimized using the AdamW optimizer with a learning rate of $1 \times 10^{-4}$ and weight decay of $1 \times 10^{-2}$.

For training \nickname, we adopt a global batch size of 256, distributed across GPUs with 8 samples per device. The model is trained for 30,000 optimization steps using a constant learning rate of $2 \times 10^{-5}$. To enhance robustness, we apply spatiotemporal masking to the input videos, with masking ratios randomly sampled from the range $[0, 0.5]$ for both temporal and spatial dimensions. To strengthen the model's fine-grained action reasoning capabilities, we perform fine-grained action reasoning training on individual videos with probability 0.4, where the model exclusively predicts latent action sequences within the expert video demonstration without proceeding to subsequent latent action and robot action prediction for the robot agent.

\begin{table*}[t]
\caption{Success rate on the LIBERO benchmark. We partition the LIBERO dataset into seen and unseen tasks and report the success rates of the compared methods on each subset.}
\centering
\begin{minipage}{0.99\textwidth}
			
			\makeatletter\def\@captype{table}
\begin{subtable}[t]{\textwidth}
\resizebox{\linewidth}{!}{
{
\fontsize{8}{10}\selectfont
\begin{tabular}{m{1.60cm}<{\centering} *{8}{m{1.20cm}<{\centering}}}
\toprule[1.5pt]
\footnotesize

Methods& DP  & OpenVLA & UniVLA & AWDA & AWDA$_R$ & \textbf{Ours} & \textbf{Ours}$_R$                   \\
\midrule
Seen & $0.70$ & $0.75$ & $0.95$ & $0.66$ & $0.62$ & $\bm{0.96}$ & $\bm{0.95}$ \\
Unseen & $0.01$ & $0.04$ & $0.13$ & $0.35$ & $0.28$ & $\bm{0.65}$ & $\bm{0.63}$ \\
\bottomrule
\end{tabular}
}
}
\end{subtable}

\makeatletter\def\@captype{table}
\begin{subtable}[t]{\textwidth}
\centering
\newcommand{\myfontsize}{\scriptsize}
\resizebox{\linewidth}{!}{
{\myfontsize
\begin{tabular}{>{\centering\arraybackslash}m{2.5cm} *{8}{>{\centering\arraybackslash}m{0.6cm}}}
    \toprule[1pt]
    & \multicolumn{2}{c}{LIBERO-Spatial} & \multicolumn{2}{c}{LIBERO-Object} & \multicolumn{2}{c}{LIBERO-Goal} & \multicolumn{2}{c}{LIBERO-Long} \\
    \cmidrule(lr){2-3} \cmidrule(lr){4-5} \cmidrule(lr){6-7} \cmidrule(lr){8-9}
    Methods & Seen & Unseen & Seen & Unseen & Seen & Unseen & Seen & Unseen \\
    \midrule
    Diffusion Policy~\cite{chi2023diffusion} & 0.76 & 0.01 & 0.90 & 0.01 & 0.68 & 0.00 & 0.45 & 0.00 \\
    OpenVLA~\cite{openvla} & 0.82 & 0.05 & 0.88 & 0.07 & 0.76 & 0.02 & 0.55 & 0.01 \\
    UniVLA \cite{bu2025univla} & 0.95 & 0.16 & 0.96 & 0.23 & 0.95 & 0.07 & 0.92 & 0.05 \\
    \hline
    \multicolumn{9}{c}{\textit{Learning from videos with the same embodiment}} \\
    \hline
    AWDA~\cite{chang2023one} & 0.71 & 0.40 & 0.78 & 0.50 & 0.63 & 0.28 & 0.51 & 0.20 \\
    \rowcolor{gray!25}\textbf{Ours} & $\bf 0.98$ & $\bf 0.70$ & $\bf 0.98$ & $\bf 0.74$ & $\bf 0.96$ & $\bf 0.62$ & $\bf 0.92$ & $\bf 0.54$ \\
    \hline
    \multicolumn{9}{c}{\textit{Learning from videos with the different embodiment}} \\
    \hline
    AWDA$_R$~\cite{chang2023one} & 0.69 & 0.32 & 0.74 & 0.41 & 0.57 & 0.21 & 0.49 & 0.17 \\
    \rowcolor{gray!25}\textbf{Ours}$_R$ & $\bf 0.95$ & $\bf 0.71$ & $\bf 0.98$ & $\bf 0.73$ & $\bf 0.95$ & $\bf 0.58$ & $\bf 0.92$ & $\bf 0.51$ \\
    \bottomrule[1pt]
  \end{tabular}
}
}
\end{subtable}
\end{minipage}
\label{tab::libero-results}
\end{table*}

\section{Experiments}
We perform experiments to answer the following questions:
\begin{itemize}
    \item Can \nickname effectively learn unseen tasks from a single expert demonstration video, enhancing the adaptability of robotic agents?
    \item Does \nickname possess robust cross-embodiment learning capabilities that enable skill acquisition from expert demonstration videos with different robotic platforms?
    \item Can \nickname directly learn robotic skills from human demonstration videos and perform robustly in real-world robotic scenarios?
\end{itemize}
We first assess the capability for unseen task learning through a comparative evaluation of \nickname against state-of-the-art VLA baselines and one-shot imitation learning (OSIL) methods (Sec.~\ref{sec:video_bc}). We then examine \nickname's capacity for cross-embodiment skill transfer in Sec.~\ref{sec:cross_robot_video}. We further demonstrate that \nickname enables effective learning from human demonstration videos and achieves robust task execution in real-world environments (Sec.~\ref{sec:human_video}). Finally, we present comprehensive ablation studies in Sec.~\ref{sec:ablation} to systematically validate the contribution of each component within our framework.

\subsection{Unseen Task Learning} \label{sec:video_bc}
We first investigate the capacity of \nickname to learn unseen tasks from videos with the same embodiment.

\myindent \textbf{Baselines}. \nickname is compared with four representative methods: 
(1) Diffusion Policy \cite{chi2023diffusion}, which represents robot visuomotor policies as conditional denoising diffusion processes, enabling stable training and natural handling of multimodal action distributions in high-dimensional spaces.
(2) AWDA  \cite{chang2023one}, which achieves one-shot visual imitation by predicting attributed waypoints from demonstration videos, executing them via hand-crafted motor primitives. 
(3) OpenVLA \cite{kim2024openvla}, a VLA based on Prismatic7B \cite{karamcheti2024prismatic} and trained on the OXE \cite{o2024open} dataset, was post-trained on the LIBERO benchmark.
(4) UniVLA \cite{bu2025univla}, which learns task-centric latent actions from diverse cross-embodiment videos without requiring action labels, pre-training the VLA model on the action-less dataset.

\begin{table*}[t]
\caption{The real-world task learning experiment results with human videos. We report the success rates of comparison methods on seen tasks and unseen tasks. Seen tasks are indicated in \colorbox{green!15}{green}, and unseen tasks are indicated in \colorbox{red!15}{red}.}
\centering
\begin{minipage}{0.99\textwidth}
			
			\makeatletter\def\@captype{table}
\begin{subtable}[t]{\textwidth}
\resizebox{\linewidth}{!}{
{
\fontsize{8}{10}\selectfont
\begin{tabular}{m{1.60cm}<{\centering} *{6}{m{2.00cm}<{\centering}}}
\toprule[1.5pt]
\footnotesize

Methods& DP  & OpenVLA & UniVLA & AWDA & \textbf{Ours}                    \\
\midrule
Seen & $0.64$ & $0.76$ & $0.86$ & $0.74$ & $\bm{0.96}$ \\
Unseen & $0.00$ & $0.04$ & $0.10$ & $0.36$ & $\bm{0.74}$ \\
\bottomrule
\end{tabular}
}
}
\end{subtable}

\makeatletter\def\@captype{table}
\begin{subtable}[t]{\textwidth}
\centering
\newcommand{\myfontsize}{\scriptsize}
\resizebox{\linewidth}{!}{
\myfontsize
\begin{tabular}{m{2.50cm}<{\centering} *{5}{m{1.55cm}<{\centering}}}
    \toprule
    \rowcolor{green!15} \cellcolor{white} {Methods} &\thead{\myfontsize Flip\\block} & 
    \thead{\myfontsize Wipe\\the tray} & 
    \thead{\myfontsize Close\\basket} & 
    \thead{\myfontsize Place fruit \\ on plate} & 
    \thead{\myfontsize Stir \\ in tray}\\
                        \midrule
                    Diffusion Policy~\cite{chi2023diffusion}  & 0.7 & 0.6 & 0.5 & 0.8 & 0.6  \\
    OpenVLA~\cite{openvla}  & 0.6 & 0.8 & 0.8 & 0.9 & 0.7  \\  
    UniVLA \cite{bu2025univla}  & 0.9 &  0.9 & 0.8 & 0.9 & 0.8 \\ 
    AWDA~\cite{chang2023one}  & 0.8 & 0.6 & 0.7 & 0.8 & 0.8 \\ 
     \textbf{Ours} & \textbf{1.0} & \textbf{0.9} & \textbf{1.0} & \textbf{1.0} & \textbf{0.9} \\ 
    
    \midrule
        
    \rowcolor{red!15} \cellcolor{white} Methods  &  \thead{\myfontsize Stack\\ block} & 
    \thead{\myfontsize Place car\\on basket} & 
    \thead{\myfontsize Beat \\ the drum} & 
    \thead{\myfontsize Push\\the toy} & 
    \thead{\myfontsize Pour from\\bowl to plate}  \\
                        
        \midrule
        Diffusion Policy~\cite{chi2023diffusion}  & 0.0 & 0.0 & 0.0 & 0.0 & 0.0  \\
    OpenVLA~\cite{openvla}  & 0.1 & 0.1 & 0.0 & 0.0 & 0.0  \\  
    UniVLA \cite{bu2025univla}  & 0.2 &  0.3 & 0.0 & 0.0 & 0.0 \\ 
    AWDA~\cite{chang2023one}  & 0.3 & 0.4 & 0.3 & 0.5 & 0.3 \\ 
     \textbf{Ours} & \textbf{0.8} & \textbf{0.8} & \textbf{0.6} & \textbf{0.7} & \textbf{0.8} \\ 
    \bottomrule
\end{tabular}
}
\end{subtable}
\end{minipage}
\label{tab::video-real}
\end{table*}

\myindent \textbf{Experiment setup}. 
We conduct our evaluation on the LIBERO benchmark~\cite{liu2024libero}, a comprehensive suite comprising 130 language-conditioned manipulation tasks. Following the experimental protocol established by OpenVLA~\cite{kim2024openvla}, we focus on four specialized suites: LIBERO-Spatial, LIBERO-Object, LIBERO-Goal, and LIBERO-Long. 
LIBERO-Spatial evaluates robustness to environmental configuration changes while maintaining consistent object types. LIBERO-Object assesses generalization across object variations under fixed spatial arrangements. LIBERO-Goal tests adaptability to different task objectives while preserving object categories and spatial layouts. LIBERO-Long presents the most challenging scenarios, requiring long-horizon reasoning across diverse object categories and spatial configurations.
Each suite contains 10 distinct tasks. To evaluate generalization capabilities, we designate 8 tasks per suite as seen tasks for training and reserve 2 tasks as unseen tasks for testing. The training set is constructed by aggregating demonstrations from all seen tasks into a consolidated dataset. For expert-agent pair construction, we leverage robot execution trajectories from the dataset, pairing videos of the same task performed by the robot as expert demonstrations. All baseline methods are trained on this consolidated dataset and subsequently evaluated on both seen and unseen task splits to assess in-distribution performance and generalization ability.

\myindent \textbf{Experiment results}. 
Table~\ref{tab::libero-results} presents a comprehensive performance comparison across all evaluated task suites, demonstrating that \nickname consistently outperforms baseline methods. Diffusion Policy (DP) exhibits limited generalization capabilities when applied to unseen tasks. Despite pre-training on the large-scale OXE dataset and subsequent post-training on the integrated LIBERO dataset, OpenVLA and UniVLA exhibit significant degraded performance on unseen tasks. This observation corroborates our hypothesis that current methods encounter significant difficulties in generalizing to novel tasks, typically requiring task-specific data collection and fine-tuning procedures to acquire novel capabilities.
Our approach also surpasses AWDA on both seen and unseen tasks, where AWDA similarly exploits expert demonstration videos for unseen task generalization.
These results substantiate our method's capacity to effectively extract fine-grained manipulation knowledge from expert demonstration videos and effectively transfer this knowledge to novel task configurations, enabling our approach to learn novel tasks from only a single expert video. 
The substantial performance gains on unseen tasks highlight the effectiveness of our approach in learning transferable representations that extend beyond the training distribution.

\begin{figure*}[h]
        \centering
    \includegraphics[width=0.98\textwidth]{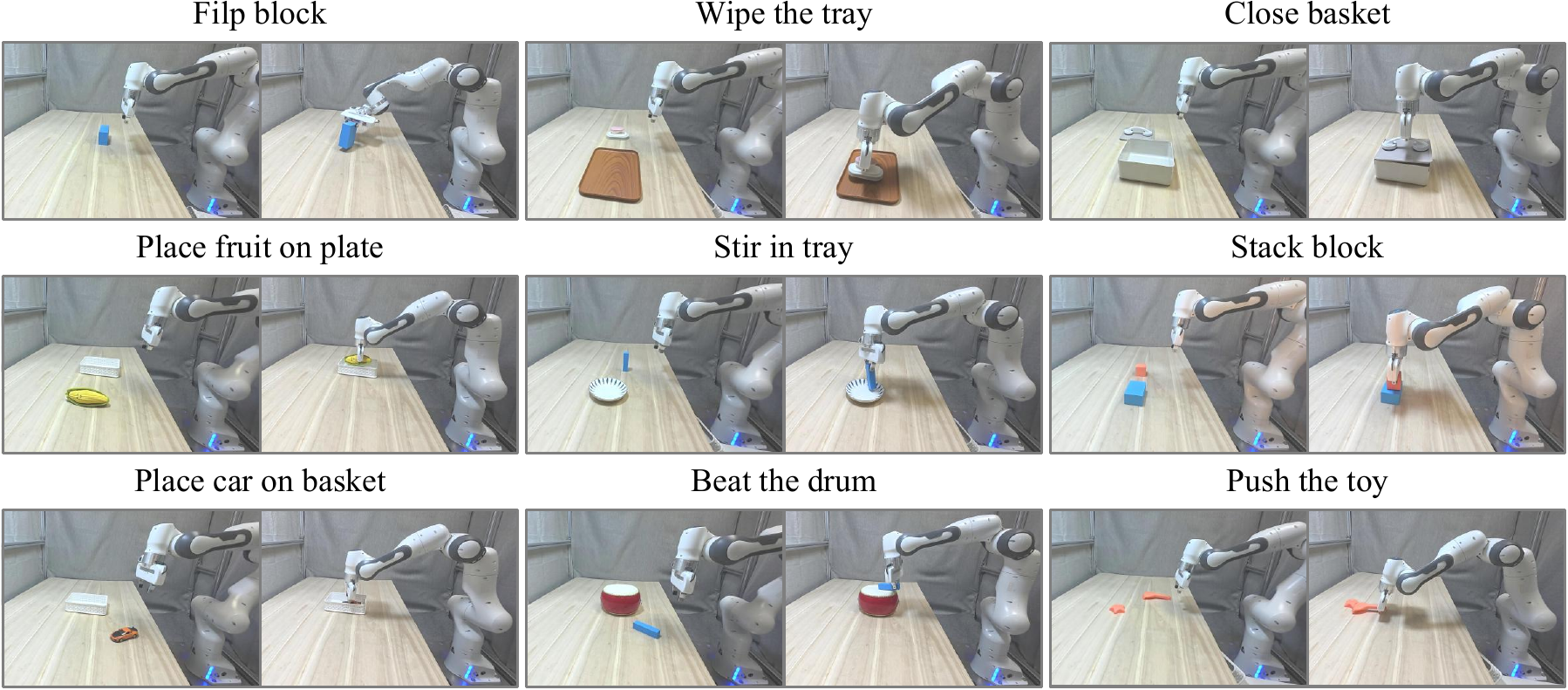}
           \caption{Example qualitative results for real-world manipulation task.}
    \label{REAL_WORLD}
\end{figure*}

\subsection{Unseen Task Learning with Cross-robot Videos} \label{sec:cross_robot_video}
We then investigate the ability of \nickname to learn novel skills from expert demonstration videos of manipulation tasks performed by a different robot platform.

\myindent\textbf{Experiment setup}. The unseen task learning experiments with cross-robot videos are conducted on the LIBERO benchmark~\cite{liu2024libero}. To construct the expert-agent pair dataset, we replay the LIBERO dataset using a UR robotic arm, where the Franka arm serves as the agent and the UR arm provides expert demonstrations. The model is post-trained on this pair dataset. During inference, the model takes demonstration videos of the UR arm performing manipulation tasks as input and generates corresponding control actions for the Franka arm to execute these tasks. All other experimental settings remain consistent with those employed in Sec.~\ref{sec:video_bc}.

\myindent\textbf{Results}. 
The empirical results for baseline approaches and the proposed method are reported in Table \ref{tab::libero-results}. Despite utilizing expert demonstration videos from heterogeneous robotic platforms, our method achieves superior performance on both seen and unseen tasks across all experimental suites. Notably, on unseen tasks, our approach exhibits strong skill acquisition capabilities by learning new skills from merely a single demonstration video at test time and executing them robustly.
The use of cross-embodiment videos incurs only marginal performance degradation compared to demonstrations from the same robotic platform. This robustness can be attributed to our latent action learning framework with cycle consistency, which effectively learns a unified latent action space. Such a design enables our method to extract generalizable manipulation knowledge from demonstrations across diverse embodiments, thereby facilitating effective manipulation knowledge extraction from cross-robot videos.

\subsection{Unseen Task Learning with Human Videos} \label{sec:human_video}

\begin{figure}[t]
    \centering
    \includegraphics[width=\linewidth]{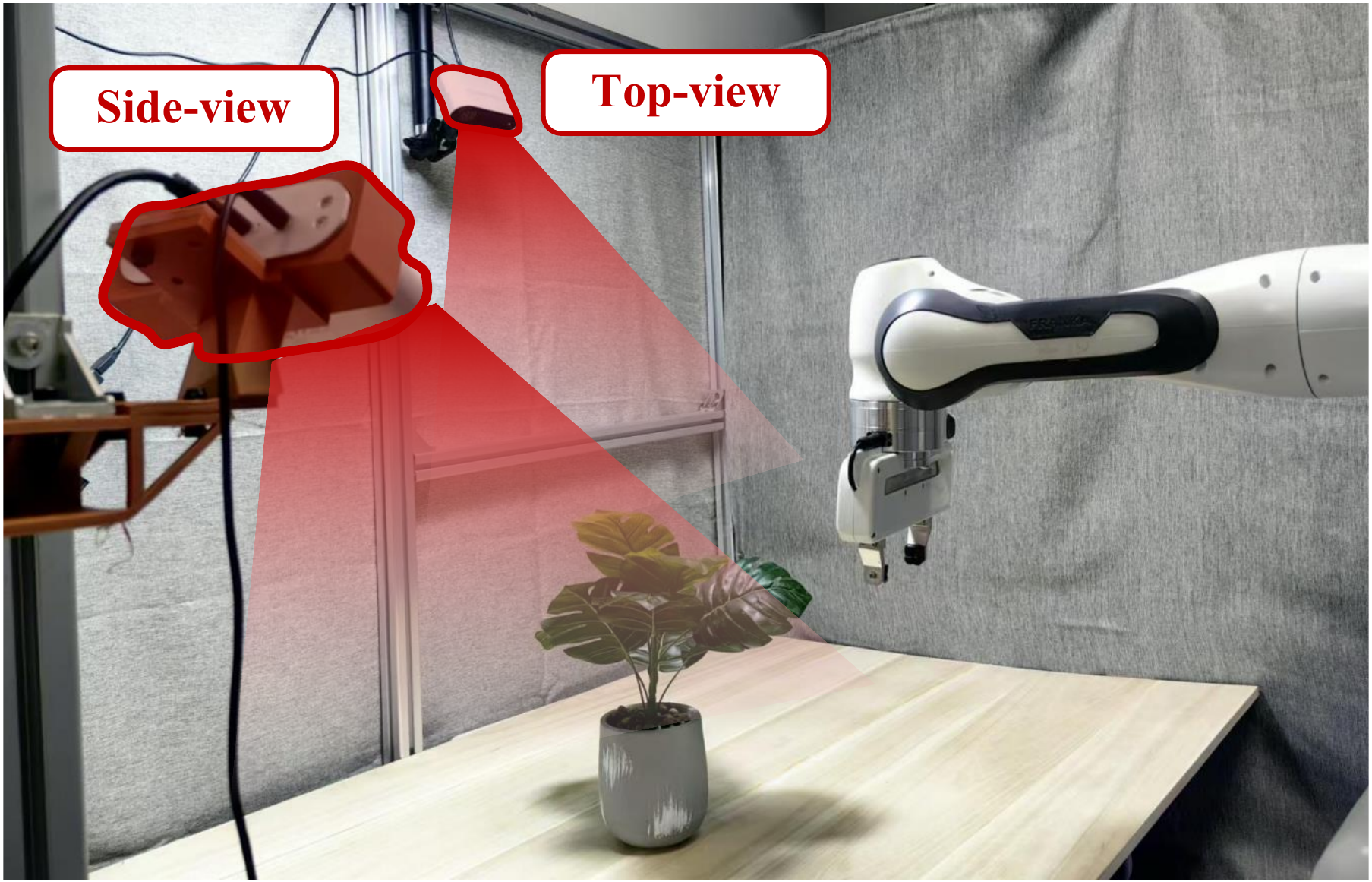}
    \caption{Real-world experiment setup.}
    \label{fig:real}
\vspace{-8pt}
\end{figure}

\myindent\textbf{Experiment setup}. 
To validate our model's capability to acquire novel skills from human demonstration videos, we train the model on our \dataname dataset, where humans provide expert demonstrations and Franka robots serve as learning agents to construct the expert-agent pair data.
We evaluate \nickname on 12 real-world manipulation tasks, consisting of 6 seen tasks and 6 unseen tasks. During evaluation, the model receives human demonstration videos corresponding to each task as input, and generates control actions for the Franka arm to execute these tasks. The real-world experimental setup is illustrated in Fig. \ref{fig:real}, featuring a seven-degree-of-freedom Franka Emika robot arm~\cite{haddadin2022franka}. Task success is determined through human evaluation, with success rates computed over 10 trials featuring randomized object positions and orientations.

\begin{table*}[th!]
\caption{{Success rates on environments with different object counts and spatial distribution. $vo$ denotes the variants evaluated in environments with different object quantities and spatial distribution.}}
\label{table::object_number}
\resizebox{\linewidth}{!}{
{\footnotesize
\begin{tabular}{>{\centering\arraybackslash}m{2cm}  *{6}{>{\centering\arraybackslash}m{1.8cm}}*{1}{>{\centering\arraybackslash}m{1.8cm}}}
\toprule
Methods &\cellcolor{green!15}\thead{  Close \\ basket} & \cellcolor{green!15}\thead{ Stir \\ in tray} & \cellcolor{red!15}\thead{ Stack \\ block} & \cellcolor{red!15}\thead{Beat \\ the drum} & \cellcolor{red!15}\thead{ Push \\ the toy}  & \thead{Overall } \\
\midrule
{Ours$\rm_{vo}$}& \textbf{0.8} & {0.7} & {0.7} & {0.6} & {0.7} & ${0.70}$   \\
\rowcolor{gray!25} {Ours}& {1.0} & {0.9} & {0.8} & {0.6} & {0.7} & ${0.80}$   \\
\bottomrule
\end{tabular}
}
}
\end{table*}

\begin{figure*}[th!]
\setlength{\abovecaptionskip}{-0.13cm}
\begin{center}
\includegraphics[width=0.98\textwidth]{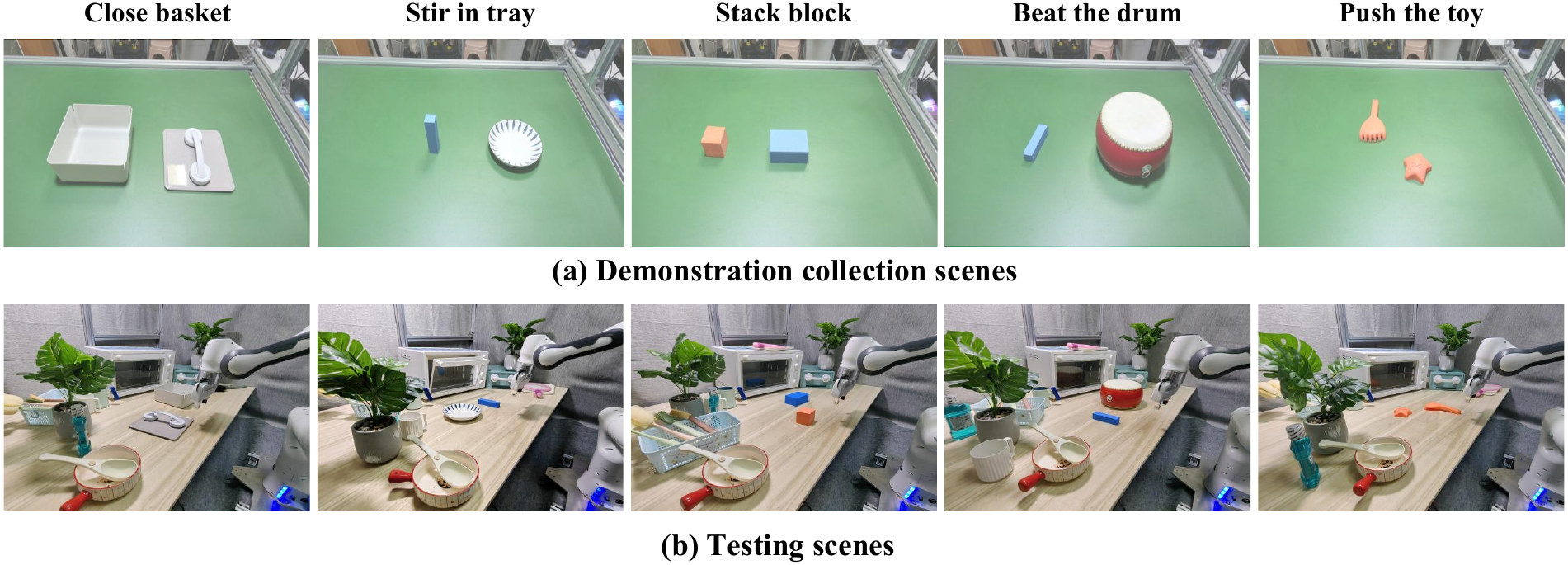}
\end{center}
\caption{{Robustness analysis on object count and spatial distribution variations.}  Seen tasks are indicated in \colorbox{green!15}{green}, and unseen tasks are indicated in \colorbox{red!15}{red}.}
\label{objects}
\end{figure*}

\begin{figure*}[t!]
    \setlength{\abovecaptionskip}{-0.13cm}
    \begin{center}
        \includegraphics[width=0.98\textwidth]{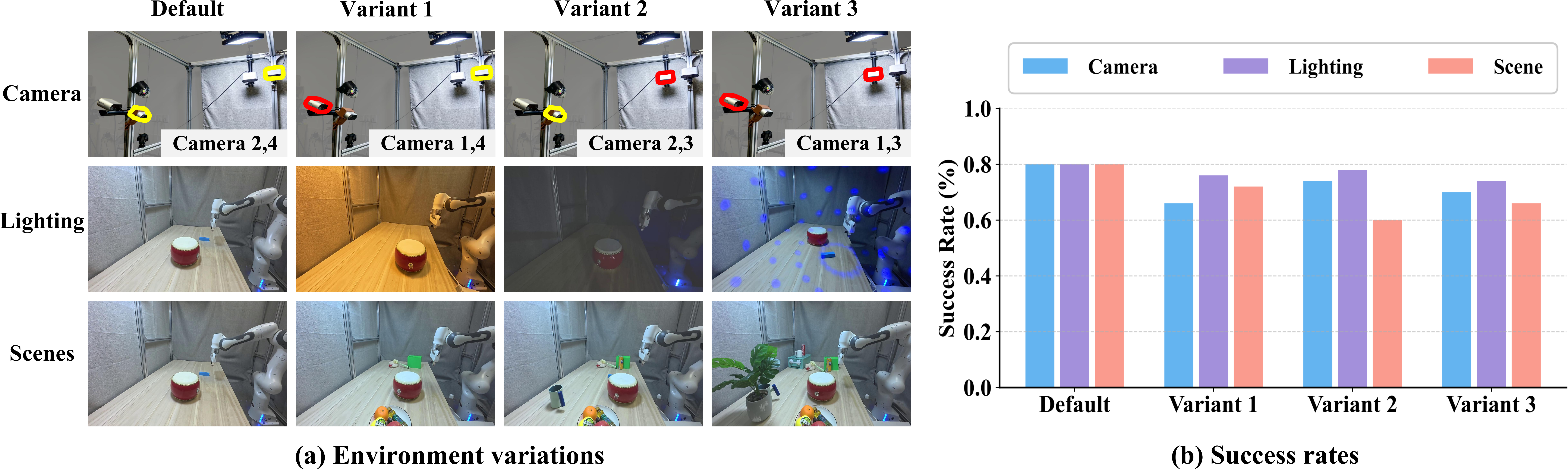}
    \end{center}
    \caption{{Environmental variations in terms of camera viewpoints, lighting, and scenes, the varying camera viewpoints are highlighted in red. Our \nickname exhibits strong robustness against these variations.}  }
    \label{Diversity}
    \end{figure*}

\myindent\textbf{Results}. 
The qualitative results are presented in Fig. \ref{REAL_WORLD}, with quantitative analysis detailed in Table~\ref{tab::video-real}. The experimental results demonstrate that \nickname achieves markedly superior performance compared to baseline methods. Our model achieves high success rates on seen tasks, validating the high fidelity of our generated robot data from the expert-agent pair data generation pipeline.
On unseen tasks, our method demonstrates promising performance, indicating its capability to effectively extract manipulation knowledge from human demonstration videos and transfer it to the robotic agent in real-world scenarios. These results substantiate that \nickname effectively bridges the embodiment gap between human demonstrations and robotic systems through its latent action learning framework. The learned unified latent action space enables effective knowledge transfer from human videos to the robot agent, facilitating efficient acquisition of new manipulation skills.
Collectively, these findings demonstrate that \nickname can learn new skills from a single human demonstration video, providing a practical approach for end-users to teach robots novel tasks.

\begin{table*}[t]
  \centering
  \caption{Ablation studies with \nickname on task learning with cross-robot videos. Default settings are marked in \colorbox{gray!25}{gray}. A3C denotes our action-centric cycle consistency.}
\resizebox{0.99\linewidth}{!}{  
\begin{tabular}{>{\centering\arraybackslash}m{4.5cm} *{8}{>{\centering\arraybackslash}m{1.2cm}}}
    \toprule[1pt]
    & \multicolumn{2}{c}{LIBERO-Spatial} & \multicolumn{2}{c}{LIBERO-Object} & \multicolumn{2}{c}{LIBERO-Goal} & \multicolumn{2}{c}{LIBERO-Long} \\
    \cmidrule(lr){2-3} \cmidrule(lr){4-5} \cmidrule(lr){6-7} \cmidrule(lr){8-9}
    Methods & Seen & Unseen & Seen & Unseen & Seen & Unseen & Seen & Unseen \\
    \midrule
    \multicolumn{9}{c}{\textit{(a) Latent action prediction}} \\
    \midrule
    W/o prediction & 0.91 & 0.48 & 0.89 & 0.55 & 0.81 & 0.47 & 0.76 & 0.33 \\
    Genie & 0.93 & 0.65 & 0.91 & 0.62 & 0.87 & 0.53 & 0.82 & 0.41 \\
        \rowcolor{gray!25}A3C & 0.95 & 0.71 & 0.98 & 0.73 & 0.95 & 0.58 & 0.90 & 0.51 \\
    \midrule
    \multicolumn{9}{c}{\textit{(b) Latent action learning framework designs}} \\
    \midrule
    W/o discriminator & 0.89 & 0.54 & 0.91 & 0.59 & 0.86 & 0.45 & 0.81 & 0.38 \\
    Local discriminator & 0.93 & 0.68 & 0.96 & 0.69 & 0.94 & 0.55 & 0.87 & 0.46 \\
    W/o Latent action buffer & 0.92 & 0.66 & 0.97 & 0.70 & 0.91 & 0.50 & 0.86 & 0.46 \\
    \rowcolor{gray!25}A3C & 0.95 & 0.71 & 0.98 & 0.73 & 0.95 & 0.58 & 0.90 & 0.51 \\
    \midrule
    \multicolumn{9}{c}{\textit{(c) Temporal-spatial masking strategy}} \\
    \midrule
    W/o masking& 0.94 & 0.64 & 0.98 & 0.67 & 0.94 & 0.54 & 0.91 & 0.46 \\
    Spatial masking & 0.95 & 0.69 & 0.97 & 0.72 & 0.95 & 0.56 & 0.90 & 0.48 \\
    \rowcolor{gray!25}Temporal-spatial masking & 0.95 & 0.71 & 0.98 & 0.73 & 0.95 & 0.58 & 0.90 & 0.51 \\
    \midrule
    \multicolumn{9}{c}{\textit{(d) Parallel modeling}} \\
    \midrule
    Auto-regressive & 0.92 & 0.63 & 0.96 & 0.67 & 0.93 & 0.52 & 0.87 & 0.35\\
    \rowcolor{gray!25}Parallel modeling & 0.95 & 0.71 & 0.98 & 0.73 & 0.95 & 0.58 & 0.90 & 0.51\\
    \bottomrule[1pt]
  \end{tabular}
}
\label{tab::Ablation}
\end{table*}

\begin{table*}[t]
\caption{{Ablation analysis on language instructions and expert video demonstrations.}}
\label{table::lang}
\resizebox{\linewidth}{!}{
{\footnotesize
\begin{tabular}{>{\centering\arraybackslash}m{2cm}  *{6}{>{\centering\arraybackslash}m{1.8cm}}*{1}{>{\centering\arraybackslash}m{1.8cm}}}
\toprule
Methods &\cellcolor{green!15}\thead{  Close \\ basket} & \cellcolor{green!15}\thead{ Stir \\ in tray} & \cellcolor{red!15}\thead{ Stack \\ block} & \cellcolor{red!15}\thead{Beat \\ the drum} & \cellcolor{red!15}\thead{ Push \\ the toy}  & \thead{Overall } \\
\midrule
{Ours$\rm_{w/o\,lang}$}& {1.0} & {0.9} & {0.7} & {0.6} & {0.6} & ${0.76}$   \\
{Ours$\rm_{w/o\,video}$}& {0.9} & {0.9} & {0.3} & {0.1} & {0.0} & ${0.44}$   \\
\rowcolor{gray!25} {Ours}& {1.0} & {0.9} & {0.8} & {0.6} & {0.7} & ${0.80}$   \\
\bottomrule
\end{tabular}
}
}
\end{table*}

\subsection{Robustness Analysis}

\textbf{Object count and spatial arrangement}.
We explore the model's robustness to variations in object quantities and spatial distributions between expert demonstration videos and agent robot manipulation scenarios. Experiments are conducted on five representative tasks: \textcolor{\mycolor}{close basket, stir in tray, stack block, beat the drum, push the toy, where the first two constitute seen tasks and the latter three represent unseen tasks}. Human demonstration videos are collected in the scenarios illustrated in Fig. \ref{objects} (a), while evaluation is performed in the scenarios depicted in Fig. \ref{objects} (b).
The experimental results presented in Table \ref{table::object_number} demonstrate that \nickname maintains high success rates and exhibits robust performance despite variations in object quantities and spatial distributions between training and evaluation scenarios. This robustness validates the effectiveness of our approach in generalizing across diverse environmental configurations.

\textbf{Environment generalization}.
We explore the model's robustness to environmental variations between expert demonstration videos and agent robot scenarios, including changes in camera perspectives, lighting conditions, and scene settings. For each factor, we design three experimental variants. Validation is conducted across five representative manipulation tasks: \textcolor{\mycolor}{close basket, stir in tray, stack block, beat the drum, push the toy}, where the first two constitute seen tasks and the latter three represent unseen tasks.
As illustrated in Fig. \ref{Diversity}, our method exhibits remarkable stability under varying lighting conditions, with minimal performance degradation. While camera perspective and scene variations exert a more pronounced impact on performance, our approach nevertheless maintains considerable robustness across different environmental settings. These results collectively demonstrate that our method achieves consistent performance across diverse environmental conditions, underscoring its adaptability and generalization capabilities.

\subsection{Ablation Analysis} \label{sec:ablation}
To investigate the fundamental designs of our \nickname approach, we perform comprehensive ablation experiments. These design decisions are evaluated in unseen task learning experiments with cross-robot videos, with performance quantified via success rate metrics.

\myindent\textbf{Latent action prediction}.  
We investigate the role of latent action prediction in learning from video demonstrations. Removing latent action prediction for both expert videos and agent robots results in significant performance degradation. This observation suggests that the latent action prediction pretraining task enhances the model's capability to recognize fine-grained actions in videos while simultaneously constructing a unified latent action space that bridges the embodiment gap between experts and agents, thereby improving the efficiency of novel skill learning from the expert video.
Compared to baseline methods that employ Genie~\cite{bruce2024genie} for latent action learning, a widely adopted approach, our method achieves substantial performance improvements. This comparison validates the critical role of cycle consistency in learning effective latent action representations. The experimental results confirm that our latent action learning framework with cycle consistency captures semantically meaningful latent actions, leading to enhanced performance.

\myindent\textbf{Latent action learning framework designs}. 
We further investigate the contributions of key components in our latent action learning framework with cycle consistency, specifically the latent action buffer and the discriminator. The experimental results are presented in Table \ref{tab::Ablation}(b).
Removing the discriminator leads to substantial performance degradation, indicating that without discriminator supervision, the decoder can leak information about the sampled latent action to the encoder, thereby undermining the effectiveness of our latent action learning approach. Additionally, employing a commonly used local discriminator that supervises individual patches still results in a performance decline. This validates the effectiveness of our local-global discriminator design, which effectively prevents information leakage and mitigates distribution mismatch between generated and dataset video frames, thus enhancing the efficiency of latent action learning.
We also examine the role of the latent action buffer. The variant that excludes the buffer and instead directly uses latent actions learned from the current batch, exhibits significant performance degradation. This decline can be attributed to the collapse of the latent action space, demonstrating that maintaining a dynamic buffer of historical latent actions is essential for stable and effective latent action learning.

\myindent\textbf{Temporal-spatial masking strategy}.
We investigate the impact of the temporal-spatial masking strategy on model performance. As shown in Table \ref{tab::Ablation}(c), removing masking entirely preserves performance on seen tasks while causing degradation on unseen tasks. This effect can be attributed to the fact that seen tasks are present in the training data, allowing the model to leverage inherent manipulation knowledge to complete these tasks even with weakened video understanding capabilities. In contrast, unseen tasks require the model to extract fine-grained manipulation knowledge directly from expert demonstration videos. The absence of the masking strategy reduces the effectiveness of latent action prediction pretraining in fostering robust video understanding capabilities, thereby impairing performance on novel tasks.
The variant employing only spatial masking also exhibits performance degradation, though less severe than complete removal. This indicates that our temporal-spatial masking approach creates a more challenging pretraining objective that requires the model to predict actions from partially visible expert demonstrations through holistic spatiotemporal video comprehension.

\myindent\textbf{Parallel modeling}. 
We investigate different action modeling approaches in Table \ref{tab::Ablation}(d). Adopting autoregressive modeling results in performance degradation on seen tasks and a decline on unseen tasks. This suggests that autoregressive modeling is susceptible to shortcut learning, which undermines the effectiveness of latent action prediction pretraining in developing the model's capability to comprehend fine-grained manipulation. Furthermore, we observe that the performance degradation is particularly pronounced on unseen tasks within the long suite compared to other task suites. This observation indicates that autoregressive modeling poses greater challenges for long-horizon latent action prediction tasks, where the accumulation of prediction errors across extended sequences exacerbates performance decline.

\textbf{Language exclusion}. 
We validate the model's semantic understanding capability for video demonstrations. We provide only expert videos and agent observations as input while removing language instructions, thereby requiring the model to infer task semantics solely from visual demonstrations. Experiments are conducted across five representative manipulation tasks: \textcolor{\mycolor}{close basket, stir in tray, stack block, beat the drum, push the toy}, where human videos serve as expert demonstrations. The experimental results presented in Table \ref{table::lang} show that our method maintains robust task completion even without language instructions, exhibiting only marginal performance degradation. This demonstrates that \nickname preserves strong video semantic understanding capabilities while developing fine-grained action recognition abilities. The model's capacity to comprehend task objectives from visual demonstrations alone highlights the effectiveness of our approach in learning rich, semantically meaningful representations from expert videos.

\textbf{Video exclusion}. 
We further validate the model's capability to operate using language instructions alone, without relying on expert video demonstrations. In this experiment, we provide only language instructions and agent observations as input while excluding expert demonstration videos.
The experimental results presented in Table \ref{table::lang} reveal a clear distinction between seen and unseen tasks. For seen tasks, \nickname maintains robust performance with language instructions alone, achieving promising success rates that indicate effective internalization of learned skills during training. In contrast, performance on unseen tasks degrades significantly without video demonstrations, with success rates dropping dramatically. This performance gap underscores our method's capacity to extract and learn novel behavioral patterns from video demonstrations, a capability that proves essential for generalizing to previously unseen manipulation tasks.

\section{Limitations}

During generalization to unseen tasks, the predominant failure cases arise from errors in basic manipulation operations, particularly imprecise grasping and inaccurate placement. We attribute these failures to two primary factors: First, under specific spatial arrangements of objects, occlusions may impede the static camera's ability to capture fine-grained robot-object interaction details. Second, since static cameras predominantly capture background regions, the proportion of task-relevant visual information input to the model remains limited. In contrast, the model can compensate for these perceptual limitations on seen tasks by leveraging its internalized manipulation knowledge acquired during training. We posit that incorporating a wrist-mounted camera represents a direct solution to this challenge, as it would enable the model to consistently observe detailed robot-object manipulations from an egocentric perspective while substantially increasing the density of task-relevant visual information in the input stream.

Additionally, we identify two methodological aspects warranting improvement: (\uppercase\expandafter{\romannumeral1})
\textcolor{\mycolor}{Our experiments reveal that the model exhibits promising error recovery capabilities during task execution. However, these capabilities can be systematically strengthened through targeted data augmentation. Specifically, during the agent demonstration generation stage of our data pipeline, we can introduce controlled trajectory perturbations paired with corresponding recovery sequences. By exposing the model to diverse failure scenarios and their corrections during training, we can enable it to learn robust error recovery strategies directly from data, thereby improving task execution reliability in real-world deployment.}
(\uppercase\expandafter{\romannumeral2})
\textcolor{\mycolor}{To maintain demonstration quality, our current pipeline leverages manually collected human videos. A promising direction for future work is to leverage internet-scale human videos to automatically generate expert-agent pair data. This would require developing a robust pipeline for video filtering, task identification, and quality assessment, but could significantly expand the diversity and volume of training data available for learning generalizable manipulation policies.}

\section{Conclusion}
This study presents \nickname, a generalist policy learning architecture that enables efficient skill acquisition from single-demonstration observation without requiring subsequent fine-tuning. Our approach processes expert demonstration videos, robot agent observations, and language instructions to predict both the latent actions exhibited in expert demonstrations and the subsequent latent actions executed by the agent.
To push the performance limit of our proposed \nickname, we develop a scalable expert-agent pair data generation pipeline capable of synthesizing paired trajectories from easily accessible videos, further augmented by curated pairs from open source datasets.
Through this pipeline, we compile a large-scale dataset comprising 892,911 expert-agent paired trajectories spanning diverse manipulation tasks. 
Extensive experiments demonstrate that our proposed \nickname is able to learn unseen tasks from a single expert demonstration at inference time. 
Our approach achieves more than 30\% improvement on unseen tasks in the LIBERO benchmark, and attains exceeding 35\% improvement when leveraging videos from different embodiments. Furthermore, our approach effectively distills knowledge from human videos, demonstrating a gain of over 38\% on real-world unseen tasks.

{
\small
\bibliographystyle{Mystyle}
\bibliography{main}
}

\newpage

\vfill

\end{document}